\documentclass[11pt]{article}

\usepackage[final]{acl}
\usepackage{xurl}

\usepackage{times}
\usepackage[T1]{fontenc}
\usepackage[utf8]{inputenc}
\usepackage{microtype}
\usepackage{inconsolata}

\usepackage{algorithm}
\usepackage{algorithmicx}
\usepackage{algpseudocode}

\usepackage{amsmath}
\usepackage{amssymb}
\usepackage{amsfonts}

\usepackage{booktabs}
\usepackage{multirow}
\usepackage{makecell}
\usepackage{subcaption}

\usepackage{xcolor}
\usepackage{colortbl}
\usepackage{pifont}

\usepackage{tcolorbox}
\tcbuselibrary{skins}

\usepackage{enumitem}

\usepackage{siunitx}
\usepackage{graphicx}
\graphicspath{{figures/}}

\colorlet{SubtotalGray}{black!8}

\definecolor{darksalmon}{rgb}{0.91, 0.59, 0.48}
\definecolor{green(pigment)}{rgb}{0.0, 0.65, 0.31}

\newcommand{\XSolidBrush}{\ding{55}}
\newcommand{\Checkmark}{\ding{51}}

\title{StructSynth: Dependency Graphs as Generation Plans for Low-Data Tabular Synthesis with Language Models}

\author{
  Siyi Liu$^{1}$ \quad
  Yujia Zheng$^{2}$ \quad
  Haoyang Li$^{3}$ \quad
  Yongqi Zhang$^{1}$\thanks{Corresponding author.} \\
  {\normalfont\normalsize $^{1}$The Hong Kong University of Science and Technology (Guangzhou)} \\
  {\normalfont\normalsize $^{2}$Carnegie Mellon University \quad
  $^{3}$The Hong Kong Polytechnic University} \\[-1pt]
  {\normalfont\normalsize\ttfamily ssui.liu1022@gmail.com \quad
  yujiazh@cmu.edu} \\
  {\normalfont\normalsize\ttfamily haoyang-comp.li@polyu.edu.hk \quad
  yongqizhang@hkust-gz.edu.cn}
}

\begin{document}

\maketitle

\begin{abstract}
Tabular data derives its value from inter-feature dependencies, yet preserving them during synthesis is fragile when samples are scarce. Existing approaches either learn dependencies implicitly through distribution fitting, rely on statistical graph learning that becomes unstable with few samples, or encode structure through flat text serialization. Recent graph-aware methods use dependency graphs as attention biases or as backbones for non-LLM samplers, but they do not use the graph as a prompt-level plan for organizing the LLM's own generation process. We introduce \textsf{StructSynth}, a framework that treats a dependency graph as a \textit{generation plan}---determining the generation order, conditioning context, and scope of each black-box LLM call. In \textbf{Evidence-Grounded Graph Induction}, LLM reasoning and statistical association cues jointly construct a Directed Acyclic Graph (DAG) from limited samples. In \textbf{Graph-Planned Conditional Synthesis}, this DAG drives autoregressive synthesis in topological order, conditioning each feature on previously generated values with graph-specified parent structure guiding each step, making the conditioning schedule explicit throughout synthesis. Experiments show that \textsf{StructSynth} achieves state-of-the-art downstream utility and the best privacy-risk ranking among fourteen compared generators in low-data settings.
\end{abstract}

\section{Introduction}

\begin{figure}[t]
\centering
\includegraphics[width=\columnwidth]{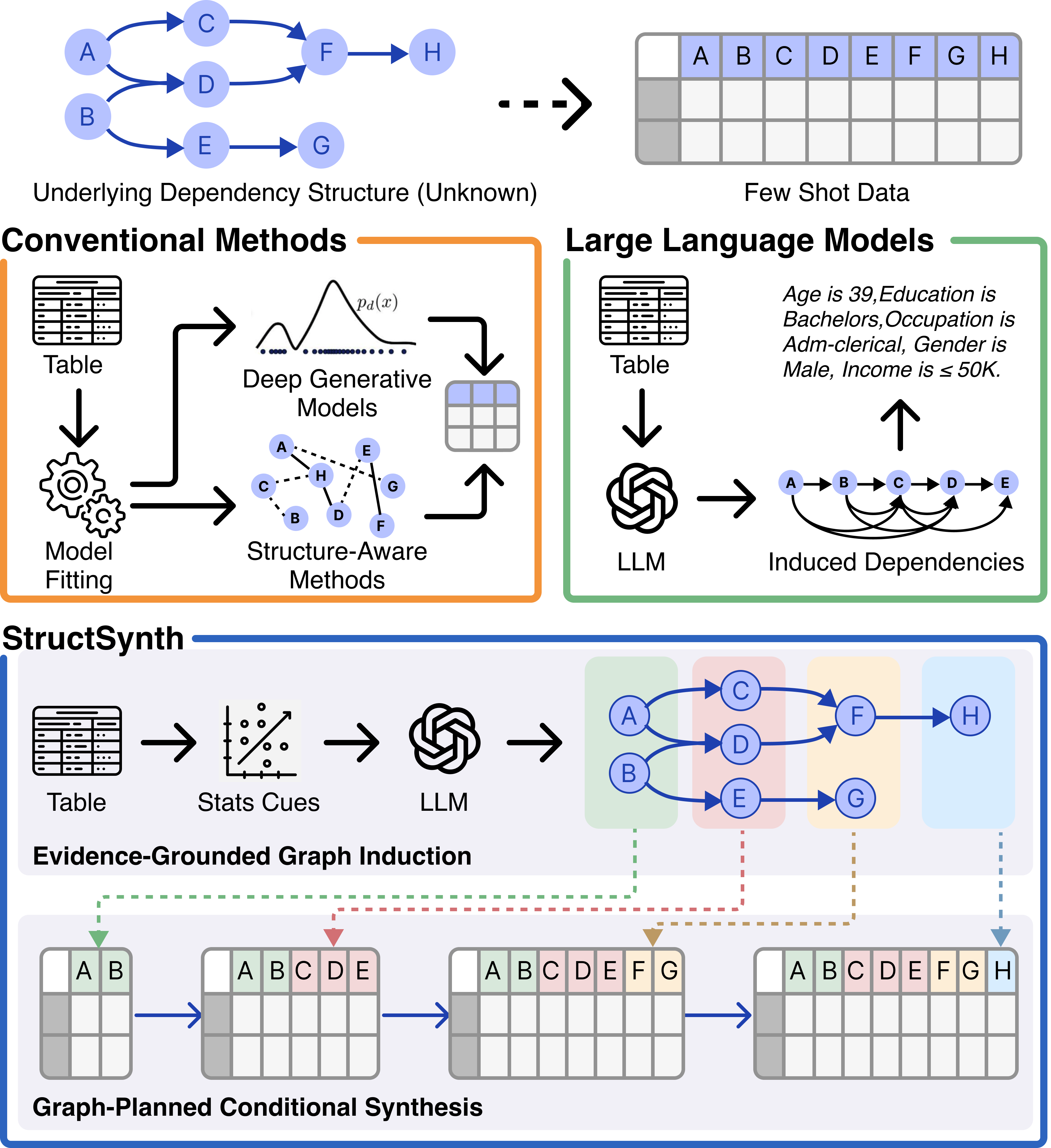}
\caption{\textbf{Tabular synthesis paradigms under data scarcity.} Conventional methods learn dependencies implicitly via distribution fitting or through learned graphical structures; LLM-based methods encode them through serialized text. \textsf{StructSynth} discovers a dependency graph and treats it as a generation plan dictating each LLM call's order, context, and scope.}

\label{fig:teaser}
\end{figure}

Low-data tabular synthesis---generating realistic records from scarce samples in domains such as healthcare~\cite{choi2017generating}, finance~\cite{rundo2019machine}, and education~\cite{luan2021review}---poses several intertwined challenges~\cite{borisov2022deep}.
Among them, \textbf{preserving inter-feature dependencies} is particularly fragile under sample scarcity: the dependency relationships that enable Machine Learning (ML) models to learn meaningful decision boundaries~\cite{fonseca2023tabular} rely on statistical signals that degrade rapidly as sample size shrinks.
Without faithful dependencies, additional synthetic samples add volume without adding predictive signal~\cite{liu2023goggle}.\looseness=-1

As illustrated in Figure~\ref{fig:teaser}, existing tabular generators handle dependencies in three important but partial ways.
\textbf{Deep generative models} capture the joint distribution end-to-end, implicitly learning inter-feature dependencies through distribution fitting; however, their performance varies across architectures, with many requiring substantially more samples to converge reliably.
\textbf{Structure-aware methods} make dependencies explicit through graphical models such as Bayesian networks~\cite{ankan2015pgmpy} and graph-based VAEs~\cite{liu2023goggle}, yet the quality of the learned graph degrades under scarcity---errors in the discovered structure propagate directly into generation~\cite{chai2022data}.
\textbf{Large Language Models (LLMs)} exploit attribute names, descriptions, and in-context examples as semantic priors~\cite{borisov2023language, cllm2024}, but most LLM-based synthesizers encode dependencies only through flat textualization of attribute--value pairs, without explicitly modeling dependency directions or conditional independence structure~\cite{liu2024rethinking}.

Recent graph-aware methods surface dependencies along different axes. PAFT~\cite{xu2024why} and GraDe~\cite{zhang-etal-2025-features} incorporate functional-dependency information during language-model fine-tuning, whereas SPADA~\cite{yang2025doubling} uses a dependency graph to guide normalizing flows. None executes a directed dependency graph as an inference-time \textit{generation plan} for a black-box LLM without parameter updates.
This gap raises a natural question: \textit{can a dependency graph determine the generation order, conditioning context, and scope of each black-box LLM call?}

We answer this question with \textsf{StructSynth}, a framework for \underline{Struct}ure-aware tabular data \underline{Synth}esis that treats a dependency graph as a generation plan---one that determines how prompts are constructed and sequenced for black-box LLM synthesis.
To use a graph as such a plan, two requirements must be met: the graph must be reliably discovered from limited samples, and it must organize the LLM's generation process. This naturally leads to a two-stage design.
In \textbf{Evidence-Grounded Graph Induction}, LLM-based reasoning is integrated with statistical association cues to construct a Directed Acyclic Graph (DAG)\footnote{We use ``DAG'' to denote a directed dependency graph used as a generative structure for organizing synthesis, not a ground-truth causal claim.} from limited samples, with a cycle resolution mechanism ensuring acyclicity. Separating discovery from synthesis allows each stage to leverage complementary signals---semantic priors and statistical evidence---without being bottlenecked by a shared scarcity constraint.
In \textbf{Graph-Planned Conditional Synthesis}, this DAG serves as the generation plan: synthesis proceeds autoregressively in topological order, with each layer of features conditioned on previously generated values and the local dependency subgraph guiding parent-aware generation at every step.
Our contributions are:
\begin{itemize}[leftmargin=*]
    \item We propose to treat a dependency graph as a \textbf{generation plan} that determines the generation order, conditioning context, and scope of each LLM call for tabular synthesis. We instantiate this idea in \textsf{StructSynth}, where \textbf{Evidence-Grounded Graph Induction} fuses LLM semantic priors with statistical association cues to discover reliable structures from limited samples, and \textbf{Graph-Planned Conditional Synthesis} compiles the discovered topology into a generation plan for layer-wise conditional synthesis.
    \item Experiments on six datasets show that \textsf{StructSynth} achieves state-of-the-art \textbf{downstream utility} and the \textbf{best privacy-risk ranking} among fourteen compared generators. Ablations confirm that both stages contribute, and that the generation plan complements rather than replaces the LLM's generative capacity.
\end{itemize}

\section{Related Work}
\label{sec:related_work}

\begin{figure*}[t]
\centering
\includegraphics[width=\textwidth]{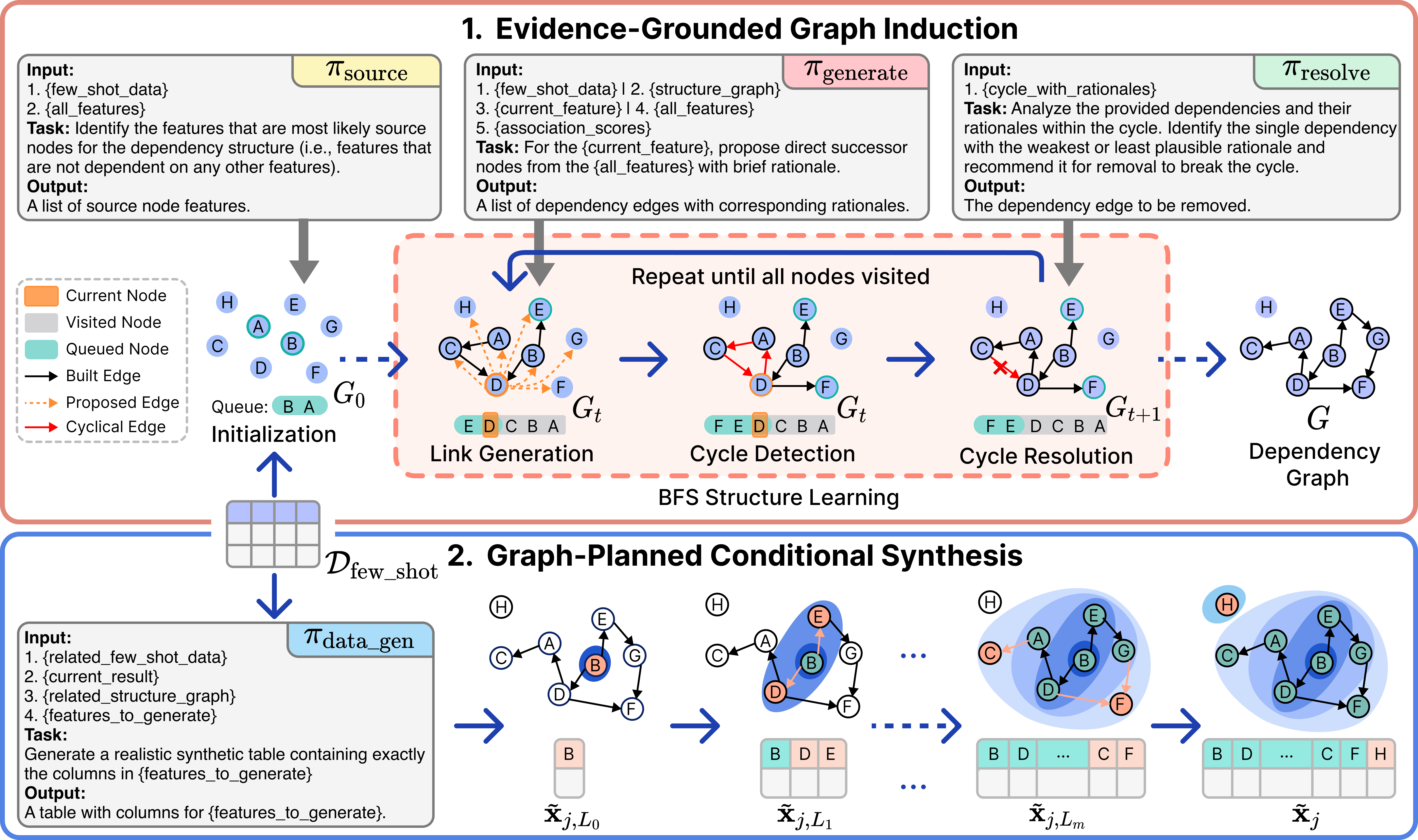}
\caption{Overview of the \textsf{StructSynth} framework. The method first performs \textbf{Evidence-Grounded Graph Induction} to infer a dependency graph from limited data via an LLM-guided search. Subsequently, \textbf{Graph-Planned Conditional Synthesis} uses the learned graph as a generation plan to autoregressively guide the LLM in generating new tabular data.}
\label{fig:pipeline}
\end{figure*}

\paragraph{Tabular Synthesis in Low-Data Regimes.}
Deep generative models---including VAE-based~\cite{patki2016synthetic}, GAN-based~\cite{xu2019modeling}, and diffusion-based~\cite{kotelnikov2023tabddpm, tabsyn} approaches---capture the joint distribution end-to-end but are data-hungry and provide limited control over inter-feature dependencies.
Structure-aware methods address this by incorporating explicit dependency graphs; Bayesian-network-based generators~\cite{ankan2015pgmpy, zhang2017privbayes} and graph-based VAEs~\cite{van2021decaf, liu2023goggle} condition synthesis on learned graphical structures, yet their effectiveness degrades when the underlying graphs cannot be reliably estimated from scarce samples.

\paragraph{LLM-Based Tabular Generation.}
Fine-tuning approaches such as GReaT~\cite{borisov2023language} and TAPTAP~\cite{zhang2023generative} serialize rows as text and train LLMs for autoregressive generation, but are computationally expensive and degrade under data scarcity.
Prompt-based methods avoid parameter updates, relying instead on inference-time prompts or example selection: Curated LLM (CLLM)~\cite{cllm2024} prompts an LLM with the available real rows and then curates the \emph{generated} samples using learning-dynamics and uncertainty signals, but does not explicitly model inter-feature dependency structure; TabGen-ICL~\cite{fang-etal-2025-tabgen} iteratively refines in-context selection to align the generated distribution with the real one.
PAFT~\cite{xu2024why} and GraDe~\cite{zhang-etal-2025-features} are structure-aware fine-tuning methods. PAFT compiles discovered functional dependencies into feature orders for permutation-aware fine-tuning, while GraDe combines FD-guided dynamic sparse attention with an FD-alignment objective during GPT-2 fine-tuning. Both tightly couple structure discovery with the generation model itself. \textsf{StructSynth} instead treats the graph as an inference-time \textit{generation plan} for a black-box LLM, without parameter updates.
SPADA~\cite{yang2025doubling} further decouples the two stages by replacing LLM-based generation with lightweight statistical estimators (KDE, normalizing flows), greatly accelerating sampling but relying on sufficient data to fit reliable conditional distributions.

\paragraph{LLM-Assisted Dependency Discovery.}
LLMs have been applied to structure learning itself. Early work demonstrated that LLMs can accurately judge pairwise causal directions from variable semantics alone~\cite{kiciman2023causal}. To scale beyond pairwise queries, BFS-based strategies reframe structure discovery as a graph traversal task, substantially reducing the number of LLM calls~\cite{jiralerspong2024efficient, zanna2025fairness}. A complementary line integrates LLM judgments with classical constraint-based algorithms, using the LLM to validate or correct statistically proposed edges~\cite{ban2023causal}.
However, these methods target causal graph identification as an end in itself and do not connect the discovered structure to a downstream data generation process.
An extended discussion of all the above lines of work is provided in Appendix~\ref{app:extended_related_work}.

\section{Methodology}

Synthesizing realistic tabular data from scarce samples requires preserving the dependency relationships among features---the very relationships that enable downstream models to learn meaningful decision boundaries.
Given a small tabular dataset $\mathcal{D}_{\mathtt{train}} = (\mathbf{X}_{\mathtt{train}}, \mathbf{A})$ with $n$ instances and $K$ attributes, where $n$ is small enough that statistical estimation of inter-feature dependencies is unreliable ($n = 100$ in our main setting; we evaluate $n \in \{20, 50, 100, 200\}$ in \S\ref{sec:shd_eval} and \S\ref{sec:influence_n}), our goal is to use an LLM-based generator $f_{\mathtt{LLM}}$ to produce a synthetic dataset $\mathcal{D}_{\mathtt{synth}}$ that, when combined with $\mathcal{D}_{\mathtt{train}}$, improves downstream model performance.

We address this by representing inter-feature dependencies as a Directed Acyclic Graph (DAG) and using it as a generation plan that organizes the LLM's synthesis process. A DAG naturally encodes generation order: its topological sort yields the synthesis sequence, and each node's parents define the conditioning set (see Appendix~\ref{sec:why_dag} for further justification). Concretely, the graph determines three aspects of the generation plan: topological layers fix the generation order (\S\ref{sec:synth}), parent sets define the conditioning context, and local subgraphs bound the scope of each prompt. As depicted in Figure~\ref{fig:pipeline}, \textsf{StructSynth} realizes this idea through two coupled stages.
\textbf{Evidence-Grounded Graph Induction} constructs the DAG from limited data by combining LLM-based reasoning with statistical association cues---the LLM's semantic prior acts as a structural regularizer while the statistical scores provide empirical grounding.
\textbf{Graph-Planned Conditional Synthesis} then uses the learned DAG as a generation plan: synthesis proceeds autoregressively in topological order, conditioning each feature on previously generated values while the local dependency subgraph makes the planned parent structure explicit throughout generation.\footnote{See Appendix~\ref{app:prompts} for full prompt templates and Algorithm~\ref{alg:structsynth} in Appendix~\ref{sec:algorithm} for the complete procedure.}

\subsection{Evidence-Grounded Graph Induction}

The dependency structure is represented as a DAG $G = (V, E)$, where the node set $V \subseteq \mathbf{A}$ contains selected attributes and edges $E$ encode probabilistic dependencies. The graph is constructed iteratively via BFS as described below.

\subsubsection{Graph Initialization}
The graph construction begins by identifying source nodes---root attributes in the generation plan that are generated without conditioning on other attributes. Using a small subset $\mathcal{D}_{\mathtt{few\_shot}} \subseteq \mathcal{D}_{\mathtt{train}}$ as in-context examples, the LLM determines the initial source nodes:
\begin{equation}
    V_{\mathtt{source}} = f_{\mathtt{LLM}}(\pi_{\mathtt{source}} ( \mathcal{D}_{\mathtt{few\_shot}})) .
\end{equation}
The graph is initialized as $G_0 = (V_{\mathtt{source}}, \emptyset)$ and expanded via BFS from these source nodes.

\subsubsection{Expansion and Reasoned Link Generation}

At iteration $t$, let $G_t = (V_t, E_t)$ denote the current graph. Each BFS step dequeues a node $A_i$ and computes normalized association scores $\mathcal{S}(A_i) \in [0,1]^{K-1}$ between $A_i$ and all other attributes~\cite{agresti2011categorical}, using Cram\'{e}r's $V$ for categorical pairs, $|r|$ (absolute Pearson correlation) for continuous pairs, and the correlation ratio for mixed pairs (see Appendix~\ref{app:association_scores} for details).
The LLM is then queried with a prompt $\pi_{\mathtt{generate}}$ that integrates the current graph state $G_t$, the node $A_i$, and $\mathcal{S}(A_i)$---pairing statistical evidence with the LLM's semantic prior so that each compensates the other's weakness at small $n$. The model proposes a set of direct successors for $A_i$, each accompanied by a textual rationale~\cite{liu-etal-2024-unleashing-power}:
\begin{equation}
    P_i = f_{\mathtt{LLM}}(\pi_{\mathtt{generate}}(A_i, G_t, \mathcal{S}(A_i), \mathcal{D}_{\mathtt{few\_shot}})).
\end{equation}
From $P_i$, new directed edges and previously unseen nodes are extracted and added to the graph and BFS queue, respectively.

\subsubsection{LLM-based Cycle Resolution}

Integrating the proposed edges into $G_t$ may introduce cycles. The LLM proposals yield a candidate edge set $E_{\mathtt{prop},t}$ and a set of newly discovered attributes $V_{\mathtt{new},t}$. A cycle detection algorithm identifies all elementary cycles $\Psi$ in $E_t \cup E_{\mathtt{prop},t}$. For each cycle $\psi \in \Psi$, the LLM is queried with $\pi_{\mathtt{resolve}}$, which presents the conflicting edges together with their rationales. The model identifies the single least-justified edge for removal; the union of all removed edges forms $E_{\mathtt{pruned}}$. The graph is then updated:
\begin{equation}
    G_{t+1} = (V_t \cup V_{\mathtt{new}, t},\ (E_t \cup E_{\mathtt{prop}, t}) \setminus E_{\mathtt{pruned}}).
\end{equation}
This expansion-resolution cycle continues until the BFS queue is empty, yielding the final DAG $G$.

\begin{table*}[ht]

    \centering
    \resizebox{\textwidth}{!}{%
    \small
    \begin{tabular}{ll llllll cr}
    \toprule
    
    \textbf{Type} & \textbf{Method} & 
    \multicolumn{1}{c}{\textbf{Adult (C)}} & 
    \multicolumn{1}{c}{\textbf{Anxiety (C)$^\dag$}} & 
    \multicolumn{1}{c}{\textbf{Compas (C)}} & 
    \multicolumn{1}{c}{\textbf{Salary (R)$^\dag$}} & 
    \multicolumn{1}{c}{\textbf{Obesity (R)}} & 
    \multicolumn{1}{c}{\textbf{Churn (C)}} & 
    \textbf{Average} & \textbf{Avg. Rank} \\
    \midrule
    \multirow{1}{*}{Real} & $\mathcal{D}_\mathtt{train}$ & $82.51_{\pm 1.93}$ & $85.49_{\pm 1.60}$ & $64.84_{\pm 2.51}$ & $49.71_{\pm 7.02}$ & $59.83_{\pm 6.40}$ & $86.86_{\pm 3.78}$ & $71.54$ & - \\
    
    \midrule
    \multirow{5}{*}{DGMs} & TVAE & $81.21_{\pm 1.54}$ & $79.63_{\pm 2.65}$ & $64.39_{\pm 2.06}$ & $46.18_{\pm 5.91}$ & \underline{$60.92_{\pm 2.61}$} & $87.42_{\pm 3.80}$ & $69.96$ & $7.17$ \\
    & DDPM & $81.72_{\pm 1.79}$ & $82.94_{\pm 3.24}$ & $66.01_{\pm 2.50}$ & $22.30_{\pm 15.70}$ & $43.20_{\pm 7.98}$ & $85.91_{\pm 2.05}$ & $63.68$ & $9.00$ \\
    & CTGAN & $81.14_{\pm 1.81}$ & $76.06_{\pm 1.47}$ & $65.17_{\pm 2.61}$ & $46.17_{\pm 5.25}$ & $54.29_{\pm 5.90}$ & $86.50_{\pm 4.17}$ & $68.22$ & $8.50$ \\
    & NFlow & $75.42_{\pm 4.58}$ & $74.83_{\pm 2.94}$ & $63.68_{\pm 4.22}$ & $25.55_{\pm 6.16}$ & $46.80_{\pm 4.17}$ & $84.99_{\pm 2.73}$ & $61.88$ & $12.17$ \\
    & TabSyn & $83.88_{\pm 1.56}$ & \underline{$85.68_{\pm 2.40}$} & \underline{$67.77_{\pm 6.98}$} & $47.17_{\pm 9.58}$ & $43.50_{\pm 9.91}$ & $89.86_{\pm 1.53}$ & $69.64$ & $4.17$ \\
    \midrule
    \multirow{4}{*}{\makecell[l]{Structure\\Aware}} & GOGGLE & $78.92_{\pm 3.04}$ & $80.92_{\pm 1.61}$ & $63.73_{\pm 3.85}$ & $43.18_{\pm 10.65}$ & $55.84_{\pm 7.77}$ & $85.27_{\pm 3.98}$ & $67.98$ & $9.33$ \\
    & BN & $82.68_{\pm 1.35}$ & $82.77_{\pm 2.39}$ & $60.13_{\pm 3.46}$ & $46.33_{\pm 7.29}$ & $59.76_{\pm 3.46}$ & $88.85_{\pm 3.02}$ & $70.09$ & $6.33$ \\
    & DECAF & $77.64_{\pm 3.91}$ & $79.90_{\pm 2.30}$ & $63.90_{\pm 3.69}$ & $43.58_{\pm 6.25}$ & $52.88_{\pm 6.44}$ & $83.94_{\pm 2.97}$ & $66.97$ & $10.17$ \\
    & SPADA-NF & $81.25_{\pm 1.35}$ & $65.39_{\pm 2.64}$ & $65.58_{\pm 3.02}$ & $35.70_{\pm 5.70}$ & $45.73_{\pm 4.52}$ & $87.55_{\pm 1.24}$ & $63.53$ & $9.33$ \\
    \midrule
    \multirow{4}{*}{LLMs} & GReaT & $82.55_{\pm 2.20}$ & $83.81_{\pm 2.55}$ & $64.02_{\pm 2.72}$ & $51.43_{\pm 3.31}$ & $47.03_{\pm 9.16}$ & $86.67_{\pm 3.20}$ & $69.25$ & $6.50$ \\
    & CLLM & \underline{$83.95_{\pm 2.09}$} & $85.17_{\pm 0.94}$ & $66.24_{\pm 1.51}$ & \underline{$54.53_{\pm 0.20}$} & $60.43_{\pm 7.19}$ & $89.81_{\pm 2.78}$ & \underline{$73.36$} & \underline{$2.83$} \\
    & GraDe & $83.72_{\pm 1.44}$ & $79.74_{\pm 2.01}$ & $66.15_{\pm 4.90}$ & $35.03_{\pm 8.08}$ & $22.92_{\pm 7.01}$ & $88.81_{\pm 2.51}$ & $62.73$ & $8.17$ \\
    & PAFT & $78.70_{\pm 2.41}$ & $72.17_{\pm 1.38}$ & $60.07_{\pm 5.48}$ & $44.76_{\pm 6.49}$ & $36.01_{\pm 7.54}$ & \underline{$89.88_{\pm 1.86}$} & $63.60$ & $10.33$ \\
    \midrule
    \multirow{1}{*}{Ours} & StructSynth & $\boldsymbol{85.55_{\pm 1.08}}$ & $\boldsymbol{86.45_{\pm 0.57}}$ & $\boldsymbol{69.40_{\pm 0.25}}$ & $\boldsymbol{55.98_{\pm 2.96}}$ & $\boldsymbol{62.27_{\pm 5.40}}$ & $\boldsymbol{90.39_{\pm 1.96}}$ & $\boldsymbol{75.01}$ & $\boldsymbol{1.00}$ \\
    
    \bottomrule
    \end{tabular}}%
    \caption{Comparison of Models on Downstream Model Performance (\%). (C) and (R) denote classification and regression tasks, respectively. $^\dag$ indicates datasets created post LLM knowledge cutoff. Best results are in \textbf{bold}; second-best are \underline{underlined}.}
    \label{tab:results_comparison_performance}
\end{table*}

\subsection{Graph-Planned Conditional Synthesis}
\label{sec:synth}

Upon learning the dependency structure $G=(V, E)$, we utilize it to guide synthetic tabular data generation. Constructing each synthetic data point $\tilde{\mathbf{x}}_j$ is performed in two sequential phases aligned with the underlying data structure.

\subsubsection{Generating Graph-Based Values}

The nodes $V$ are partitioned into topological layers $\mathcal{L} = (L_1, \dots, L_m)$. For each layer $L_i$, we extract the dependency subgraph $G_i = G[L_i \cup \Gamma^{-}(L_i)]$, where $\Gamma^{-}(L_i)$ denotes parent nodes, along with the corresponding columns from the few-shot examples.
Values for layer $L_i$ are generated conditionally on all previously generated values $\tilde{\mathbf{x}}_{j, <i}$, with the local subgraph $G_i$ specifying the parent dependencies to preserve:
\begin{equation}
\tilde{\mathbf{x}}_{j,L_i} = f_{\mathtt{LLM}}\!\Big(\pi_{\mathtt{data\_gen}}\!\big(\tilde{\mathbf{x}}_{j,<i},\, G_i,\, L_i,\, \mathcal{D}_{\mathtt{few\_shot}}^i\big)\!\Big).
\end{equation}
Iterating through all layers yields a partial synthetic data point $\tilde{\mathbf{x}}_{j, V}$ comprising values for all graph-dependent attributes.

\subsubsection{Completing Synthesis}

Values for isolated attributes $A_{\mathtt{iso}} = \mathbf{A} \setminus V$ (those outside the dependency graph) are generated in a single step, conditioned on all graph-based values $\tilde{\mathbf{x}}_{j, V}$:
\begin{equation}
\tilde{\mathbf{x}}_{j,A_{\mathtt{iso}}} = f_{\mathtt{LLM}}\!\Big(\pi_{\mathtt{iso}}\!\big(\tilde{\mathbf{x}}_{j,V},\, A_{\mathtt{iso}},\, \mathcal{D}_{\mathtt{few\_shot}}^{\mathtt{iso}}\big)\!\Big).
\end{equation}
The graph-dependent and isolated values are concatenated to form each synthetic record; repeating $s$ times yields $\mathcal{D}_{\mathtt{synth}} = (\mathbf{X}_{\mathtt{synth}}, \mathbf{A})$.

\section{Experiments}
\subsection{Experimental Setups}
\paragraph{Datasets.}
We conduct experiments on six tabular datasets: Adult~\cite{adult_2}, Anxiety~\cite{kaggleSocialAnxiety}, Compas~\cite{propublicaMachineBias}, Salary~\cite{kaggleGlobalMarket}, Obesity~\cite{obesity}, and Churn~\cite{churn}. These datasets cover a range of domains (e.g., social, medical, business) and downstream tasks including binary/multi-class classification and regression. Notably, Anxiety and Salary were released after the knowledge cutoff of modern LLMs, allowing us to assess performance in the absence of prior exposure. Full dataset details are provided in Appendix~\ref{app:datasets}. Additionally, for evaluating structure discovery quality, we utilize three benchmark datasets from bnlearn~\cite{scutari2010learning} with known ground-truth DAGs: Asia~\cite{bnlearn-repo-asia} (8 nodes), Child~\cite{bnlearn-repo-child} (20 nodes), and Insurance~\cite{bnlearn-repo-insurance} (27 nodes). These allow direct measurement of structural recovery via the Structural Hamming Distance (SHD)~\cite{de2009comparison}.

\paragraph{Baselines.}
We compare our method against thirteen baselines from three categories (implementation details in Appendix~\ref{app:baselines}): 1)~\textbf{Deep Generative Models (DGMs)}, including TVAE~\cite{xu2019modeling}, CTGAN~\cite{xu2019modeling}, TabDDPM~\cite{kotelnikov2023tabddpm}, NFlow~\cite{papamakarios2021normalizing}, and TabSyn~\cite{tabsyn}; 2)~\textbf{Structure-Aware Methods}, including Bayesian Networks~\cite{ankan2015pgmpy}, GOGGLE~\cite{liu2023goggle}, DECAF~\cite{van2021decaf}, and SPADA-NF~\cite{yang2025doubling}; and 3)~\textbf{LLM-based Methods}, including GReaT~\cite{borisov2023language}, CLLM~\cite{cllm2024}, GraDe~\cite{zhang-etal-2025-features}, and PAFT~\cite{xu2024why}.
For downstream utility, the selection combines established baselines with recent representatives in each category. All methods are evaluated with $n=100$ real examples, matching the low-data setting targeted in this work.

\paragraph{Evaluation Metrics.}
Following previous work~\cite{borisov2023language, cllm2024}, we evaluate synthetic data along three axes (detailed definitions in Appendix~\ref{app:metrics}): \textbf{Downstream Model Performance} measures the utility of synthetic data by training a downstream model on $\mathcal{D}_{\mathtt{aug}} = \mathcal{D}_{\mathtt{train}} \cup \mathcal{D}_{\mathtt{synth}}$ and evaluating on $\mathcal{D}_{\mathtt{test}}$; we report AUC for classification and $R^2$ for regression. \textbf{Statistical Fidelity Error} quantifies how well pairwise correlations in the synthetic data match the real data (\textbf{lower is better}). \textbf{Privacy Risk}~\cite{van2023membership} measures whether synthetic records are overly similar to training records; the score is the fraction whose nearest neighbor is drawn from $\mathcal{D}_{\mathtt{train}}$ (\textbf{ideal: 0.5}).

\paragraph{Implementation Details.}

We use \texttt{gpt-4o-mini} with a temperature of 0.9 for both the CLLM and \textsf{StructSynth} models. Our experiments evaluate the low-data regime, generating 1000 synthetic samples ($s=1000$) from 100 original samples ($n=100$). Downstream performance is assessed using an XGBoost model. All experiments are repeated 10 times, and we report mean $\pm$ standard deviation. Unless otherwise stated, $\mathcal{D}_{\mathtt{few\_shot}} = \mathcal{D}_{\mathtt{train}}$ in the 100-shot setting; Appendix~\ref{app:hyperparameter} lists the full hyperparameter configurations.

\begin{table*}[ht]
\centering
\resizebox{\textwidth}{!}{%
\small
\begin{tabular}{ll llllll cr}
\toprule

\textbf{Type} & \textbf{Method} & 
\multicolumn{1}{c}{\textbf{Adult}} &
\multicolumn{1}{c}{\textbf{Anxiety$^\dag$}} &
\multicolumn{1}{c}{\textbf{Compas}} &
\multicolumn{1}{c}{\textbf{Salary$^\dag$}} &
\multicolumn{1}{c}{\textbf{Obesity}} &
\multicolumn{1}{c}{\textbf{Churn}} &
\textbf{Average} & \textbf{Avg. Rank} \\

\midrule
\multicolumn{10}{l}{{Results for \textbf{Privacy Preservation}.} \textit{Ranks are computed by deviation from 0.5, i.e., $|\mathrm{PrivacyRisk}-0.5|$; lower deviation indicates better privacy.}} \\
\midrule
    \multirow{5}{*}{DGMs} & TVAE & \underline{$50.34_{\pm 3.91}$} & $55.41_{\pm 4.26}$ & $64.50_{\pm 5.18}$ & $55.09_{\pm 4.38}$ & $52.93_{\pm 7.86}$ & $58.70_{\pm 4.75}$ & $56.16$ & $4.50$ \\
    & DDPM & $99.95_{\pm 0.05}$ & $66.20_{\pm 13.17}$ & $67.18_{\pm 18.06}$ & $63.32_{\pm 19.52}$ & $71.36_{\pm 16.88}$ & $62.89_{\pm 24.91}$ & $71.82$ & $10.83$ \\
    & CTGAN & $50.70_{\pm 3.64}$ & $56.64_{\pm 4.72}$ & $64.43_{\pm 4.22}$ & $53.59_{\pm 2.05}$ & $53.44_{\pm 3.89}$ & $57.84_{\pm 2.31}$ & $56.11$ & $4.83$ \\
    & NFlow & $54.36_{\pm 1.92}$ & $\boldsymbol{53.52_{\pm 3.08}}$ & $62.57_{\pm 3.66}$ & $52.76_{\pm 2.34}$ & $51.74_{\pm 3.68}$ & $52.97_{\pm 3.33}$ & $54.65$ & $3.67$ \\
    & TabSyn & $51.64_{\pm 2.67}$ & $70.53_{\pm 7.45}$ & $67.55_{\pm 5.40}$ & $54.66_{\pm 4.19}$ & $63.30_{\pm 7.19}$ & $59.38_{\pm 3.70}$ & $61.18$ & $8.00$ \\
\midrule
    \multirow{4}{*}{\makecell[l]{Structure\\Aware}} & GOGGLE & $49.15_{\pm 7.97}$ & $43.62_{\pm 33.49}$ & $65.16_{\pm 17.25}$ & $30.94_{\pm 18.49}$ & $44.28_{\pm 23.83}$ & $61.87_{\pm 18.01}$ & \underline{$49.17$} & $7.50$ \\
    & BN & $53.73_{\pm 2.36}$ & $72.98_{\pm 4.69}$ & $67.78_{\pm 3.63}$ & $61.50_{\pm 1.48}$ & $68.02_{\pm 2.18}$ & $59.68_{\pm 3.74}$ & $63.95$ & $9.50$ \\
    & DECAF & $46.23_{\pm 19.13}$ & $58.59_{\pm 10.37}$ & $64.07_{\pm 19.14}$ & $56.91_{\pm 17.41}$ & \underline{$50.82_{\pm 15.49}$} & $69.79_{\pm 18.81}$ & $57.74$ & $6.67$ \\
    & SPADA-NF & $95.20_{\pm 0.71}$ & $78.76_{\pm 15.53}$ & $\boldsymbol{57.86_{\pm 11.26}}$ & $94.98_{\pm 0.88}$ & $83.08_{\pm 1.69}$ & $92.30_{\pm 1.49}$ & $83.70$ & $11.00$ \\
    \midrule
    \multirow{4}{*}{LLMs} & GReaT & $99.50_{\pm 0.35}$ & $85.39_{\pm 2.29}$ & $94.10_{\pm 1.33}$ & $86.66_{\pm 2.45}$ & $80.75_{\pm 2.99}$ & $94.51_{\pm 1.48}$ & $90.15$ & $13.50$ \\
    & CLLM & $49.45_{\pm 3.85}$ & $44.37_{\pm 4.55}$ & $64.14_{\pm 4.22}$ & \underline{$50.96_{\pm 5.40}$} & $45.36_{\pm 2.69}$ & $\boldsymbol{48.78_{\pm 5.82}}$ & $\boldsymbol{50.51}$ & \underline{$3.50$} \\
    & GraDe & $97.51_{\pm 1.60}$ & $61.14_{\pm 1.94}$ & $75.11_{\pm 5.11}$ & $65.99_{\pm 2.60}$ & $59.30_{\pm 4.76}$ & $69.07_{\pm 2.86}$ & $71.35$ & $10.67$ \\
    & PAFT & $76.35_{\pm 3.68}$ & $60.76_{\pm 3.46}$ & $75.31_{\pm 4.56}$ & $60.28_{\pm 2.31}$ & $57.08_{\pm 6.87}$ & $62.32_{\pm 4.27}$ & $65.35$ & $9.33$ \\
    \midrule
    \multirow{1}{*}{Ours} & StructSynth & $\boldsymbol{49.97_{\pm 5.78}}$ & \underline{$44.74_{\pm 5.13}$} & \underline{$62.37_{\pm 3.86}$} & $\boldsymbol{50.13_{\pm 4.89}}$ & $\boldsymbol{49.80_{\pm 6.31}}$ & \underline{$48.67_{\pm 4.84}$} & $50.95$ & $\boldsymbol{1.50}$ \\

\midrule
\midrule
\multicolumn{10}{l}{{Results for \textbf{Statistical Fidelity Error}.} \textit{A lower score indicates higher fidelity.}} \\
\midrule
    \multirow{5}{*}{DGMs} & TVAE & $53.71_{\pm 1.80}$ & $46.16_{\pm 1.28}$ & $60.26_{\pm 1.09}$ & $66.93_{\pm 1.13}$ & $58.41_{\pm 0.52}$ & $53.22_{\pm 1.54}$ & $56.45$ & $7.33$ \\
    & DDPM & $65.26_{\pm 1.05}$ & $52.76_{\pm 1.22}$ & $64.94_{\pm 1.79}$ & $90.51_{\pm 0.33}$ & $64.44_{\pm 1.36}$ & $64.40_{\pm 2.90}$ & $67.05$ & $12.33$ \\
    & CTGAN & $52.37_{\pm 1.56}$ & $47.14_{\pm 2.39}$ & $59.96_{\pm 0.90}$ & $58.57_{\pm 0.89}$ & $58.71_{\pm 0.57}$ & $52.32_{\pm 1.75}$ & $54.85$ & $6.17$ \\
    & NFlow & $50.68_{\pm 2.71}$ & $43.50_{\pm 1.77}$ & $60.62_{\pm 0.93}$ & $\boldsymbol{52.38_{\pm 1.10}}$ & $58.85_{\pm 0.98}$ & $53.98_{\pm 2.24}$ & \underline{$53.34$} & $6.00$ \\
    & TabSyn & \underline{$49.17_{\pm 1.34}$} & $55.31_{\pm 1.31}$ & $57.23_{\pm 2.24}$ & $62.90_{\pm 3.01}$ & $58.08_{\pm 0.54}$ & $53.99_{\pm 2.08}$ & $56.11$ & $6.83$ \\
\midrule
    \multirow{4}{*}{\makecell[l]{Structure\\Aware}} & GOGGLE & $68.32_{\pm 1.38}$ & $51.58_{\pm 7.27}$ & $66.19_{\pm 1.13}$ & $87.15_{\pm 1.65}$ & $63.80_{\pm 2.40}$ & $65.51_{\pm 2.16}$ & $67.09$ & $12.33$ \\
    & BN & $\boldsymbol{47.46_{\pm 2.15}}$ & $\boldsymbol{38.57_{\pm 1.01}}$ & $60.33_{\pm 0.82}$ & \underline{$54.39_{\pm 1.52}$} & $57.42_{\pm 0.75}$ & \underline{$48.10_{\pm 1.29}$} & $\boldsymbol{51.05}$ & $\boldsymbol{3.33}$ \\
    & DECAF & $62.71_{\pm 3.21}$ & $48.53_{\pm 1.53}$ & $63.82_{\pm 1.34}$ & $84.70_{\pm 2.41}$ & $60.93_{\pm 2.21}$ & $60.22_{\pm 0.88}$ & $63.49$ & $10.50$ \\
    & SPADA-NF & $83.14_{\pm 0.73}$ & $45.92_{\pm 1.33}$ & \underline{$53.62_{\pm 1.55}$} & $61.79_{\pm 0.55}$ & $\boldsymbol{55.29_{\pm 0.85}}$ & $\boldsymbol{47.85_{\pm 1.17}}$ & $57.94$ & $4.83$ \\
    \midrule
    \multirow{4}{*}{LLMs} & GReaT & $82.71_{\pm 1.35}$ & \underline{$40.10_{\pm 1.64}$} & $\boldsymbol{52.17_{\pm 1.18}}$ & $58.75_{\pm 0.89}$ & \underline{$56.83_{\pm 0.87}$} & $48.18_{\pm 2.16}$ & $56.46$ & \underline{$4.17$} \\
    & CLLM & $61.93_{\pm 4.43}$ & $57.64_{\pm 1.51}$ & $59.74_{\pm 2.33}$ & $78.60_{\pm 0.98}$ & $65.74_{\pm 1.42}$ & $62.20_{\pm 3.68}$ & $64.31$ & $10.83$ \\
    & GraDe & $84.30_{\pm 0.86}$ & $42.59_{\pm 0.93}$ & $53.82_{\pm 2.62}$ & $55.85_{\pm 1.42}$ & $57.44_{\pm 1.43}$ & $53.66_{\pm 1.90}$ & $57.94$ & $5.67$ \\
    & PAFT & $51.66_{\pm 2.25}$ & $42.96_{\pm 2.19}$ & $53.86_{\pm 4.88}$ & $62.22_{\pm 0.87}$ & $57.00_{\pm 1.37}$ & $54.72_{\pm 2.16}$ & $53.74$ & $5.17$ \\
    \midrule
    Ours & StructSynth & $57.60_{\pm 3.13}$ & $57.86_{\pm 1.98}$ & $57.23_{\pm 1.69}$ & $64.64_{\pm 0.96}$ & $61.14_{\pm 1.10}$ & $58.63_{\pm 2.66}$ & $59.52$ & $9.50$ \\
\bottomrule
\end{tabular}}%
\caption{Comparison of Models on Privacy Preservation and Statistical Fidelity Error (\%). $^\dag$ indicates datasets created post LLM knowledge cutoff. Best results are in \textbf{bold}; second-best are \underline{underlined}.}
\label{tab:results_comparison_merged}

\end{table*}

\subsection{Main Results}
\paragraph{Downstream Model Performance.}
Table~\ref{tab:results_comparison_performance} summarizes the downstream results. \textsf{StructSynth} achieves the best mean performance on all six datasets, with a perfect average rank of 1.00 and the highest average score (75.01). This improves over using only $\mathcal{D}_\mathtt{train}$ (71.54) and surpasses the strongest competitor, CLLM (73.36), demonstrating that explicitly guiding an LLM with a graph-based generation plan yields substantial gains. Notably, these gains hold on the two post-cutoff datasets (Anxiety, Salary), where the LLM lacks prior exposure, showing that the advantage does not depend on the backbone having seen these tables during pre-training. Standard DGMs and structure-aware models perform notably weaker, as implicit distribution learning and data-driven graph discovery are unreliable in our $n=100$ low-data setting.
Separately, under the same 100-shot XGBoost evaluation, \textsf{StructSynth} also achieves higher downstream utility than the recent fine-tuning baseline PAFT~\cite{xu2024why} on all six datasets, with average scores of 75.01 and 63.60, respectively.

\paragraph{Sensitivity to Downstream Model Choice.}
Beyond the XGBoost-based main evaluation, Table~\ref{tab:downstream_sensitivity} compares downstream utility across four evaluator families---XGBoost, Random Forest, Logistic Regression/Ridge, and MLP---on four representative datasets. \textsf{StructSynth} attains the highest score in 14 of 16 dataset--evaluator combinations and leads in all 12 classification combinations. On Salary, it leads under XGBoost and Random Forest; under Ridge, $\mathcal{D}_\mathtt{train}$ alone performs best, and under MLP, CLLM slightly outperforms \textsf{StructSynth}, suggesting that the utility gain on this regression task is evaluator-dependent.

\begin{table}[t]
\small
\centering
\setlength{\tabcolsep}{0.9mm}
\begin{tabular}{ll|ccc}
    \hline
    \textbf{Dataset} & \textbf{Eval.\ Model} & $\mathcal{D}_\mathtt{train}$ & \textbf{CLLM} & \textbf{StructSynth} \\
    \hline
    \multirow{4}{*}{\makecell[l]{Adult\\(C)}}
    & XGBoost & $82.51$ & $83.95$ & $\boldsymbol{85.55}$ \\
    & Random Forest & $85.03$ & $84.63$ & $\boldsymbol{87.45}$ \\
    & Logistic Reg. & $85.68$ & $85.49$ & $\boldsymbol{87.80}$ \\
    & MLP & $83.63$ & $83.59$ & $\boldsymbol{86.85}$ \\
    \hline
    \multirow{4}{*}{\makecell[l]{Anxiety$^\dag$\\(C)}}
    & XGBoost & $85.49$ & $85.17$ & $\boldsymbol{86.45}$ \\
    & Random Forest & $86.85$ & $85.14$ & $\boldsymbol{87.04}$ \\
    & Logistic Reg. & $87.26$ & $85.87$ & $\boldsymbol{87.82}$ \\
    & MLP & $84.85$ & $82.27$ & $\boldsymbol{85.76}$ \\
    \hline
    \multirow{4}{*}{\makecell[l]{Salary$^\dag$\\(R)}}
    & XGBoost & $49.71$ & $54.53$ & $\boldsymbol{55.98}$ \\
    & Random Forest & $55.16$ & $51.29$ & $\boldsymbol{57.12}$ \\
    & Ridge & $\boldsymbol{59.48}$ & $52.58$ & $56.80$ \\
    & MLP & $44.61$ & $\boldsymbol{46.00}$ & $42.64$ \\
    \hline
    \multirow{4}{*}{\makecell[l]{Churn\\(C)}}
    & XGBoost & $86.86$ & $89.81$ & $\boldsymbol{90.39}$ \\
    & Random Forest & $90.78$ & $91.66$ & $\boldsymbol{92.49}$ \\
    & Logistic Reg. & $90.61$ & $91.91$ & $\boldsymbol{92.19}$ \\
    & MLP & $91.71$ & $92.36$ & $\boldsymbol{93.61}$ \\
    \hline
\end{tabular}
\caption{Downstream utility (\%) across four evaluator families and four representative datasets. Values are means over 10 runs. (C) denotes AUC for classification, (R) denotes $R^2$ for regression, and $^\dag$ marks a post-cutoff dataset. The highest score in each row is in \textbf{bold}.}
\label{tab:downstream_sensitivity}
\end{table}

\paragraph{Privacy Preservation and Statistical Fidelity Error.}

Table~\ref{tab:results_comparison_merged} reveals a recurring tension between fidelity and privacy: the highest-fidelity methods are often the ones that memorize. For example, BN achieves the best fidelity (avg.\ rank 3.33) but exhibits poor privacy (avg.\ rank 9.50). GReaT ranks second in fidelity (avg.\ rank 4.17) yet has the worst privacy risk (90.15\%), essentially reproducing training records. Similarly, SPADA-NF attains strong fidelity (avg.\ rank 4.83) at the cost of poor privacy (avg.\ rank 11.00). At the other end, prompt-based CLLM achieves strong privacy (avg.\ rank 3.50) but weaker fidelity (avg.\ rank 10.83). PAFT shows a similar imbalance, ranking 5.17 in statistical fidelity but 9.33 in privacy. \textsf{StructSynth} achieves the best privacy preservation (avg.\ rank 1.50) with a statistical fidelity rank of 9.50. Importantly, high pairwise fidelity does not guarantee downstream utility: the \textit{Bayesian Sampler} ablation achieves the best raw fidelity (48.86) yet the lowest AUC (81.17). This is because Statistical Fidelity measures \emph{pairwise} correlations, whereas \textsf{StructSynth}'s DAG encodes \emph{conditional} dependencies between parent and child nodes---the relationships a downstream model relies on. The generation plan thus acts as a regularizer: it constrains overfitting to individual records while preserving the conditional structure that matters most for utility.

\begin{figure*}[t!]
\centering
\includegraphics[width=\textwidth]{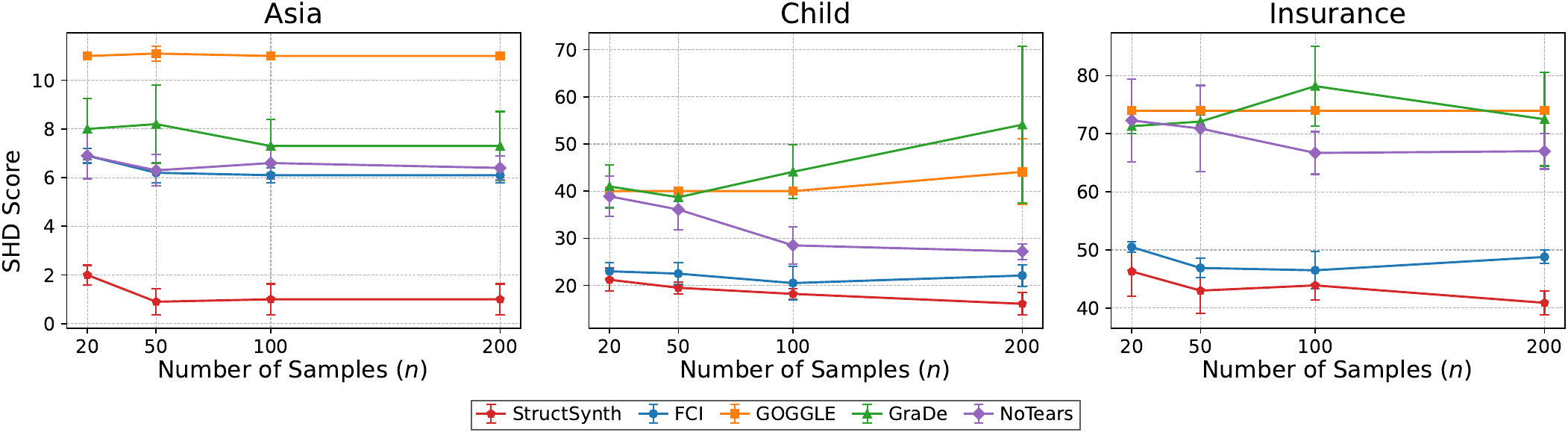}
\caption{SHD of discovered dependency graphs vs. number of training samples on Asia, Child, and Insurance datasets with ground-truth DAGs. Lower SHD indicates better structural recovery.}
\label{fig:shd}
\end{figure*}

\subsection{Ablation Study}

To isolate the contributions of our method's two main stages, we evaluate seven variants. For the structure learning stage: 1) \textit{PC Discovery}, substituting our LLM-guided discovery with the classical Peter--Clark algorithm~\cite{spirtes2000causation}; 2) \textit{NoTears Discovery}, replacing our method with the continuous optimization-based NoTears algorithm~\cite{zheng2018dags}; 3) \textit{No-Association Scores}, removing statistical association scores from the prompt to examine their guiding influence; and 4) \textit{Pairwise Discovery}, replacing the efficient BFS traversal with less scalable pairwise queries to evaluate the performance-efficiency trade-off. For the data generation stage: 5) \textit{No Topological Order}, conducting data generation based on the complete learned structure but ignoring the topological order; 6) \textit{No Structure}, removing structural guidance entirely (equivalent to the CLLM); and 7) \textit{Bayesian Sampler}, using a Bayesian sampler over the learned~graph instead of the LLM generator.

\begin{table}[t!]
\small
\centering
\setlength{\tabcolsep}{1mm}
\resizebox{\columnwidth}{!}{%
\begin{tabular}{ll|ccc}
    \hline
    \multicolumn{2}{c|}{\textbf{Method}} & \textbf{AUC} & \textbf{Fidelity} & \textbf{Privacy} \\ 
    \hline
    \multicolumn{2}{c|}{\textsf{StructSynth}}& \textbf{85.55} & 57.60 & \underline{49.97} \\
    \hline
    \multirow{4}{*}{\makecell[l]{Structure\\Learning}} & PC Discovery& 84.54 & 59.69 & 50.52 \\
    & NoTears Discovery& 84.17 & 59.49 & 47.85 \\
    & No-Association Scores& 84.06 & 59.10 & 46.60 \\
    & Pairwise Discovery & \underline{85.25} & 57.37 & 49.17 \\
    \hline
    \multirow{3}{*}{Generation} & No Topological Order & 84.41 & 60.47 & 51.03 \\
    & No Structure & 83.95 & 61.93 & 49.45 \\
    & Bayesian Sampler & 81.17 & \textbf{48.86} & \textbf{50.02} \\
    \hline
\end{tabular}
}
\caption{Ablation Study on the Adult Dataset.}
\label{tab:ablation_study}

\end{table}

The results (Table~\ref{tab:ablation_study}) yield three insights. First, \textbf{LLM-guided discovery outperforms classical alternatives}: \textsf{StructSynth} exceeds both \textit{PC} and \textit{NoTears} by 1.0--1.4 AUC pts, showing that semantic priors stabilize structure learning under data scarcity. Second, \textbf{structural guidance is multifaceted}: removing the graph (\textit{No Structure}, $-1.6$ pts) or ignoring topological ordering (\textit{No Topological Order}, $-1.1$ pts) both degrade utility. Removing the statistical association cues (\textit{No-Association Scores}, $-1.5$ pts) degrades utility nearly as much as removing the graph entirely, confirming that the evidence-grounding component---not the LLM prior alone---does substantive work in graph induction. Finally, \textbf{the plan organizes the generator rather than replacing it}: substituting a \textit{Bayesian Sampler} over the same discovered graph causes the sharpest drop ($-4.4$ pts), the largest single effect in the table. The graph contributes on top of a strong LLM generator rather than in place of it.
Taken together, these results clarify why \textsf{StructSynth}'s advantage over CLLM is not merely incremental: CLLM processes features in an arbitrary serialized order, implicitly inducing dependencies from the input sequence, whereas \textsf{StructSynth} encodes dependencies as an explicit DAG so that relationships are \emph{discovered} rather than incidentally induced.

\paragraph{Robustness to Anonymized Schemas.}
Complementing the \textit{No-Association Scores} ablation, Table~\ref{tab:anonymization} examines the other input channel by replacing all column names and task/column descriptions with generic identifiers while retaining cell values. Relative to the original schema, anonymization changes utility from 0.865 to 0.850 AUC on Anxiety and from 0.560 to 0.462 $R^2$ on Salary. Fidelity Error decreases on both datasets, while DCR (Privacy Risk) moves farther from 0.50, particularly on Salary. These results indicate that statistical association cues retain useful signal without descriptive schema semantics, while the utility impact varies by task.

\begin{table}[t]
\centering
\small
\setlength{\tabcolsep}{3.5pt}
\resizebox{\columnwidth}{!}{%
\begin{tabular}{llccc}
\toprule
\textbf{Dataset} & \textbf{Schema} & \textbf{Utility} & \textbf{Fidelity} & \textbf{Privacy} \\
\midrule
\multirow{2}{*}{Anxiety}
  & Original   & \textbf{0.865} & 0.579          & \textbf{0.447} \\
  & Anonymized & 0.850          & \textbf{0.534} & 0.564 \\
\midrule
\multirow{2}{*}{Salary}
  & Original   & \textbf{0.560} & 0.646          & \textbf{0.501} \\
  & Anonymized & 0.462          & \textbf{0.615} & 0.412 \\
\bottomrule
\end{tabular}
}
\caption{Task-and-field anonymization on two post-cutoff datasets ($n=100$, 1{,}000 synthetic samples). Utility is AUC for Anxiety and $R^2$ for Salary. Lower Fidelity is better; Privacy (DCR) is evaluated relative to 0.50. Best results are in \textbf{bold}.}
\label{tab:anonymization}
\end{table}

\begin{figure*}[t!]
\centering
\includegraphics[width=\textwidth]{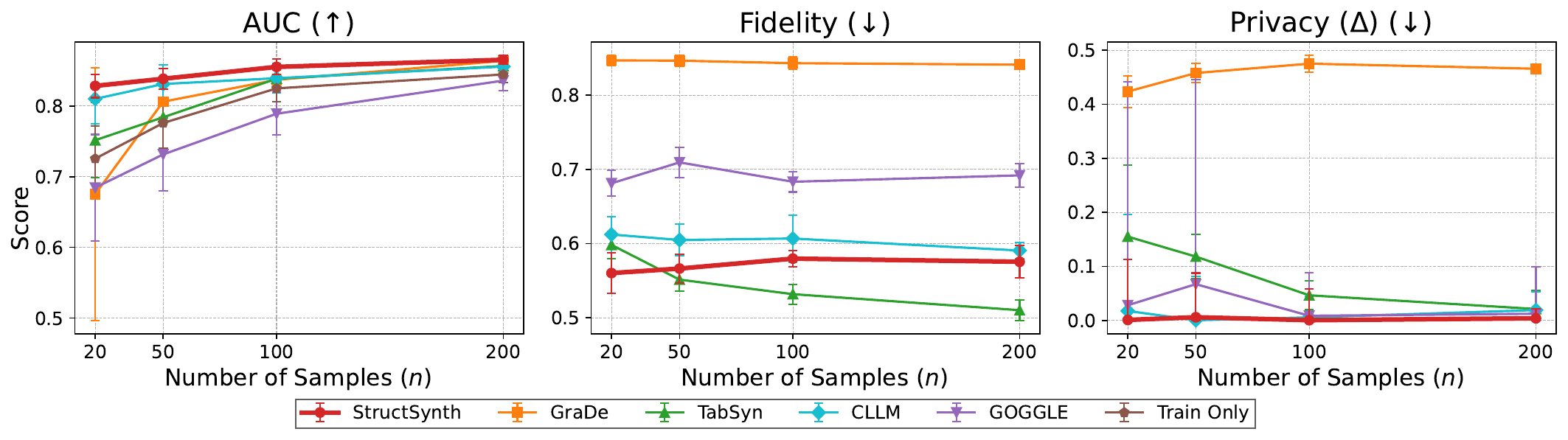}
\caption{Effect of training sample size ($n$) on downstream utility (AUC$\uparrow$), statistical fidelity ($\downarrow$), and privacy risk deviation $\Delta$ ($\downarrow$) on the Adult dataset.}
\label{fig:influence_n}
\end{figure*}

\begin{figure}[h]
\centering
\includegraphics[width=\columnwidth]{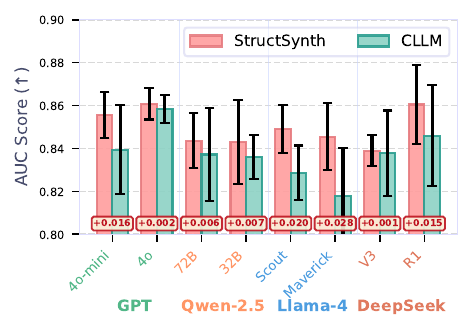}
\caption{AUC on Adult across LLM backbones.}
\label{fig:llm_compare}
\end{figure}

\subsection{Structural Fidelity under Ground-Truth Graphs}
\label{sec:shd_eval}

We evaluate structure discovery quality on three \texttt{bnlearn} benchmarks with known ground-truth DAGs---Asia (8 nodes), Child (20 nodes), and Insurance (27 nodes)---using the Structural Hamming Distance (SHD), which counts the edge insertions, deletions, or reversals needed to recover the true graph (lower is better).

Figure~\ref{fig:shd} shows SHD across varying sample sizes ($n \in \{20, 50, 100, 200\}$). \textsf{StructSynth} achieves the lowest or near-lowest SHD on all three datasets, with the advantage most pronounced in low-data settings ($n \leq 50$) where baseline graph-learning methods such as FCI, GOGGLE, and NoTears produce substantially higher errors. This indicates that the LLM's semantic prior effectively complements weak statistical signals, enabling robust structure recovery even from severely limited samples. Detailed graph visualizations are provided in Appendix~\ref{app:qualitative_graph_analysis}.

\subsection{Influence of Training Sample Size}
\label{sec:influence_n}

To evaluate data efficiency, we vary the training size $n$ from 20 to 200 on the Adult dataset and measure AUC, Statistical Fidelity Error, and Privacy Risk deviation $\Delta = |\text{PrivacyRisk} - 0.5|$ (Figure~\ref{fig:influence_n}). \textsf{StructSynth} maintains high AUC, low fidelity error, and near-zero $\Delta$ across all sample sizes. The advantage is most pronounced in the low-data regime ($n \leq 50$), where baselines like BN and GOGGLE require significantly more samples to become competitive, while \textsf{StructSynth} sustains stable performance. This suggests that the generation plan acts as a regularizer: it prevents record-level overfitting and yields a stable utility--privacy--fidelity balance across the low-data range we target.
We extend the Adult sample-size analysis to the post-cutoff Anxiety and Salary datasets (Appendix~\ref{app:post_cutoff_robustness}). Across $n\in\{20,50,100,200\}$, \textsf{StructSynth} achieves higher utility than CLLM at every evaluated size on both datasets, with the largest margin of 3.31 AUC points on Anxiety at $n=20$ (Table~\ref{tab:post_cutoff_vary_n}).

\subsection{Influence of Different Language Models}

To assess generalizability, we benchmark \textsf{StructSynth} against CLLM using a diverse set of open-source and proprietary LLMs on the Adult dataset (Figure~\ref{fig:llm_compare}). \textsf{StructSynth} outperforms CLLM across every tested architecture. The gain is largest for mid-capability models (Llama-4 Maverick: $+0.028$ AUC) and smallest for the strongest backbones, consistent with structural guidance mattering most where the model's own priors are weakest. Appendix~\ref{sec:token_usage} provides a token usage analysis showing that \textsf{StructSynth} incurs a modest +31\% overhead over CLLM.
We further evaluate Qwen2.5-32B on the two post-cutoff datasets (Table~\ref{tab:qwen25_post_cutoff}). \textsf{StructSynth} obtains higher mean utility on both: 0.856 versus 0.843 AUC on Anxiety and 0.532 versus 0.518 $R^2$ on Salary. The consistent direction provides additional evidence that the gains transfer beyond the GPT backbone, although the margins are modest relative to the reported standard deviations.

\section{Conclusion}

We presented \textsf{StructSynth}, a framework that treats a learned dependency graph as a generation plan for LLM-based tabular synthesis. Separating graph induction from data generation lets each stage leverage complementary signals---semantic priors for structure discovery and the discovered topology for conditional generation.
Experiments on six datasets show that this design achieves state-of-the-art downstream utility and the best privacy-risk ranking among the compared generators, with gains that hold across LLM backbones.
More broadly, our results suggest that \emph{explicit structural guidance} is a promising direction for tabular synthesis under data scarcity: investing in a reliable generation plan yields consistent improvements over unstructured approaches.

\section*{Limitations}

\paragraph{Evaluation range.}
Our evaluation focuses on low-data settings with $n\leq 200$; whether \textsf{StructSynth}'s advantages over baselines persist at larger sample sizes remains untested.

\paragraph{DAG assumption.}
The directed acyclic graph formulation, while well-suited for autoregressive generation, cannot directly represent cyclic or bidirectional dependencies.
For data governed by genuine feedback loops---such as reciprocal economic indicators---the framework must approximate such relationships with unidirectional edges, potentially losing part of the underlying dependency structure.

\paragraph{LLM API dependence.}
The current implementation relies on commercial LLM APIs, raising cost and reproducibility concerns.
While Appendix~\ref{sec:token_usage} reports modest token overhead and \textsf{StructSynth} outperforms CLLM on every one of the eight LLM backbones we tested, API version updates may still affect exact reproducibility of the discovered structures.

\section*{Ethical Considerations}

\paragraph{Data and privacy.}
Our experiments use publicly available tabular datasets and benchmark Bayesian networks; we do not collect new data or conduct human-subject studies.
Some evaluated datasets involve sensitive human-centered domains, so we use them only as research benchmarks.
\textsf{StructSynth} is not a formal privacy mechanism, and applying it to confidential data may expose information through prompts, logs, or generated records.
Privacy-critical uses should therefore require additional disclosure-risk assessment and appropriate access controls.

\paragraph{Bias and misuse.}
Synthetic data may preserve or amplify biases in the source data, particularly for demographic, medical, legal, or socioeconomic attributes.
Generated records should not be used for high-stakes decision making without domain review, fairness auditing, and validation on real held-out data.
When hosted LLMs are used, users should ensure that prompts and example rows comply with provider policies and applicable data-governance requirements.

\section*{Acknowledgments}
This work was supported by the Guangdong Basic and Applied Basic Research Foundation (2025A1515010304, 2026A1515010227), Guangdong Province Project 2024QN11X088, and Guangzhou Science and Technology Planning Project 2025A03J4491.

\bibliography{references}

\newpage
\appendix

\section{Code Availability}
\label{app:code_availability}

Our implementation, together with instructions for reproducing the experiments, is available at:
\url{https://github.com/ssui-liu/StructSynth}.

\section{Extended Related Work}
\label{app:extended_related_work}

This section provides an extended discussion of the related work summarized in Section~\ref{sec:related_work}, offering additional technical details on representative methods in each line of research.

\subsection{Early and Statistical Methods}
Early tabular data synthesis methods relied on statistical techniques. SMOTE~\cite{chawla2002smote} generates synthetic minority-class samples by interpolating between nearest neighbors in feature space, addressing class imbalance but limited to local linear combinations. Copula-based methods~\cite{xue2000multivariate} model the joint distribution by separately estimating marginal distributions and their dependency structure through copula functions, providing a principled statistical framework but struggling to capture complex, non-linear inter-feature relationships. These foundational approaches established the core objectives of tabular synthesis---realistic sample generation with dependency preservation---but their limited expressiveness motivated the shift toward deep generative models.

\subsection{Deep Generative Models for Tabular Data}
Deep generative models learn the data distribution through expressive neural architectures. TVAE~\cite{patki2016synthetic} adapts the Variational Autoencoder framework to mixed-type tabular data by applying mode-specific normalization to continuous columns. CTGAN~\cite{xu2019modeling} introduces a conditional GAN with a training-by-sampling strategy that addresses the imbalanced categorical distributions common in tabular datasets. Normalizing Flow-based methods~\cite{papamakarios2021normalizing} learn invertible transformations between a simple base distribution and the data distribution, offering exact likelihood computation. More recently, diffusion-based approaches have achieved strong results: TabDDPM~\cite{kotelnikov2023tabddpm} applies denoising diffusion probabilistic models to tabular data with separate handling of numerical and categorical features, while TabSyn~\cite{tabsyn} first encodes mixed-type tabular data into a continuous latent space via a VAE and then applies score-based diffusion within that space, improving inter-column relation fidelity while reducing inference steps. Despite these advances, all DGM-based approaches learn feature dependencies only as a byproduct of distribution fitting, offering limited interpretability or explicit control over the dependency structure during generation.

\subsection{Structure-Aware Generative Methods}
Structure-aware methods explicitly incorporate graphical models of feature dependencies into the generative process. Bayesian Network-based generators~\cite{ankan2015pgmpy} learn a directed acyclic graph over features and sample from the resulting factorized joint distribution. PrivBayes~\cite{zhang2017privbayes} extends this framework with differential privacy guarantees by injecting calibrated noise into the learned conditional distributions, enabling privacy-preserving synthesis. Among deep-learning approaches, DECAF~\cite{van2021decaf} leverages a user-specified or learned causal graph to achieve fairness-aware generation by intervening on sensitive attributes, while GOGGLE~\cite{liu2023goggle} employs a graph-based VAE that jointly learns the dependency structure and the generative model in an end-to-end differentiable manner. However, the quality of these methods is fundamentally tied to the accuracy of the underlying graph, which degrades substantially under limited-sample settings where conditional independence tests lack statistical power and score-based search becomes prone to local optima. More recently, SPADA~\cite{yang2025doubling} employs an LLM to annotate sparse feature dependencies as a directed graph, then synthesizes data by traversing the graph with kernel density estimation or conditional normalizing flows, bypassing LLM-based generation entirely to achieve substantial sampling speedups.

\subsection{LLM-Based Tabular Generation: Extended Discussion}
Beyond the prompt-based and structure-aware LLM methods discussed in the main text, several fine-tuning approaches have been proposed. GReaT~\cite{borisov2023language} serializes each table row as a natural-language sentence with randomized column ordering and fine-tunes a pre-trained GPT-2 model for autoregressive row generation. TAPTAP~\cite{zhang2023generative} introduces a table pre-training stage that exposes the LLM to diverse tabular datasets before task-specific fine-tuning, improving generalization across domains. AIGT~\cite{zhang2025aigt} further augments the fine-tuning paradigm with attribute-level instruction tuning, enabling the model to follow explicit generation directives for individual columns. While these fine-tuning methods can achieve high fidelity when sufficient training data is available, they incur substantial computational costs and their performance degrades markedly in low-data regimes ($n \le 100$) where the limited training set is insufficient for effective parameter adaptation.

\section{Structure Learning Background}
\label{app:structure_learning_background}

\subsection{Conventional Structure Learning}
Conventional approaches to dependency structure discovery primarily fall into two categories: constraint-based and score-based methods. Constraint-based methods, such as the Peter--Clark (PC) algorithm~\cite{spirtes2000causation} and the Fast Causal Inference (FCI) algorithm~\cite{spirtes1995causal}, utilize conditional independence tests to iteratively prune edges from an initially fully connected graph. While PC assumes causal sufficiency (no hidden confounders), FCI is specifically designed to handle latent variables. In contrast, score-based methods optimize a global scoring criterion to identify the most suitable graph structure. A representative example is the Greedy Equivalence Search (GES)~\cite{chickering2002optimal}, which searches through equivalence classes to maximize a scoring function. However, both constraint-based and score-based approaches typically yield only Markov equivalence classes, thereby failing to distinguish between structurally distinct graphs that encode identical conditional independencies. To address this limitation, Functional Causal Models (FCMs) impose stronger assumptions regarding the data generation process. FCMs model effects explicitly as functions of direct causes combined with independent noise terms, enabling the identification of causal directions since independence between cause and noise typically breaks down when reversed~\cite{shimizu2006linear, hoyer2008nonlinear, zhang2009identifiability}. Recent advancements have generalized this approach, introducing conditions such as Generalized Independent Noise (GIN) to manage latent causal variables in linear, non-Gaussian settings~\cite{xie2020generalized}. However, the statistical power of these traditional algorithms heavily relies on a large number of samples, making them unreliable for structure discovery in the low-data regimes addressed by our work.

\subsection{LLM-based Structure Learning}
Leveraging Large Language Models (LLMs) for structure learning has recently emerged as a promising direction due to their inherent reasoning capabilities. Several paradigms have been proposed for integrating LLMs into structure discovery. Initially, methods focused on querying LLMs about pairwise causal relationships, showing strong accuracy in determining causal directions between variable pairs~\cite{kiciman2023causal}. To overcome the quadratic complexity inherent to pairwise approaches, more scalable methods employing breadth-first search (BFS) strategies have been introduced, reframing structure learning as a graph traversal task and significantly reducing the required number of queries to the LLM~\cite{jiralerspong2024efficient, zanna2025fairness}. A third paradigm enhances reliability by integrating an iterative supervision loop, in which traditional structure-learning algorithms propose initial graphs that are subsequently validated or corrected by LLMs. This method grounds LLM-generated structures in empirical evidence, thereby reducing hallucination risks~\cite{ban2023causal}. LLM-based structure discovery methods have demonstrated notable effectiveness across diverse domains, including medicine~\cite{naik2024applying, antonucci2023zero, arsenyan2024large}, economics~\cite{causal4financial}, social sciences~\cite{manning2024automated}, and complex networks~\cite{mao2024critical}. Inspired by the success of LLMs in leveraging prior knowledge for domain-specific reasoning~\cite{liu2025agenthpo}, our work integrates a scalable, LLM-driven paradigm for robust structure discovery.

\section{Structural Formulation: Why a DAG?}
\label{sec:why_dag}

A central design choice in \textsf{StructSynth} is the representation of inter-feature dependencies as a Directed Acyclic Graph (DAG). Since tabular feature relationships are often associative rather than inherently directional, this choice warrants explicit justification. We adopt the DAG formulation for three principled reasons.

\paragraph{Pragmatic Dependency Structure, Not Causal Claim.}
We emphasize that our DAG is a \emph{generative dependency structure}: its edges encode statistical and probabilistic dependencies inferred from limited observations, not claims of ground-truth causality. This pragmatic interpretation aligns with the foundational Bayesian Network (BN) literature~\cite{pearl2009causality, koller2009probabilistic}, where DAGs serve as compact encodings of conditional independence relationships without necessarily asserting causal direction. The distinction is important: \textsf{StructSynth} uses the discovered DAG to \emph{organize} the generation process, not to make ontological claims about the data-generating mechanism.

\paragraph{Functional Necessity of Directionality.}
The Graph-Planned Conditional Synthesis stage (Section~\ref{sec:synth}) generates feature values autoregressively: each feature is produced conditioned on values already generated for its parent nodes. This autoregressive process inherently requires a \emph{total ordering} over features to determine which values are available as conditioning context at each step. An undirected graph $\{A\text{---}B, B\text{---}C\}$ is ambiguous---it cannot specify whether $A$ should be generated before $C$ or vice versa. A DAG $\{A \to B \to C\}$ resolves this ambiguity by defining a topological order that directly translates into a layer-by-layer generation schedule, much as directed dependency structures schedule computation in other graph-based settings~\cite{du2026graphoracle}. Even when the underlying statistical relationship between two features is symmetric (e.g., $\mathrm{Age} \leftrightarrow \mathrm{Income}$), the generation process \emph{must} select a direction---the DAG provides exactly this. The specific direction chosen does not affect \emph{whether} the dependency is captured, only the \emph{order} in which values are conditioned.

\paragraph{Acyclicity as a Mathematical Prerequisite.}
Beyond directionality, acyclicity is essential for the generation process to be well-defined. If cycles were permitted---e.g., $A \to B \to C \to A$---the generation would encounter circular dependencies: the value of $A$ would depend on $C$, which depends on $B$, which in turn depends on $A$, creating an unresolvable loop. The DAG constraint guarantees the existence of a valid topological ordering over nodes, ensuring that every feature can be generated exactly once, conditioned only on features whose values have already been determined.

\paragraph{Summary.}
In summary, the DAG in \textsf{StructSynth} serves a \emph{functional} rather than \emph{ontological} role: it imposes a well-defined, acyclic generation order that enables structure-conditioned autoregressive synthesis. This formulation is consistent with decades of probabilistic graphical model research and is a natural fit for our graph-as-plan formulation.

\section{Statistical Association Measures}
\label{sec:association_scores}
\label{app:association_scores}

To quantify the empirical dependencies between attribute pairs, the framework employs three association measures, chosen according to the data types of the variables involved.

\subsection{Continuous-Continuous: Pearson's R}
For two continuous variables $x = \{x_i\}_{i=1}^n$ and $y = \{y_i\}_{i=1}^n$, Pearson's correlation coefficient $r$ is defined as the standardized covariance:
\begin{equation}
    r = \frac{\sum_{i=1}^{n}(x_i-\bar{x})(y_i-\bar{y})}{\sqrt{\sum_{i=1}^{n}(x_i-\bar{x})^2}\sqrt{\sum_{i=1}^{n}(y_i-\bar{y})^2}}
\end{equation}
where $\bar{x}$ and $\bar{y}$ denote the sample means. The value of $r$ ranges from $[-1, 1]$, where the sign indicates the direction of the linear relationship and the absolute value indicates its strength.

\subsection{Categorical-Continuous: Correlation Ratio}
The Correlation Ratio, $\eta$, measures the relationship between a categorical variable $x$ and a continuous variable $y$. It is defined as the square root of the weighted variance of the category means divided by the total variance:
\begin{equation}
    \eta = \sqrt{\frac{\sum_{x} n_x(\bar{y}_x - \bar{y})^2}{\sum_{i=1}^{n}(y_i - \bar{y})^2}}
\end{equation}
where $n_x$ is the number of samples in category $x$, $\bar{y}_x$ is the mean of $y$ for samples in category $x$, and $\bar{y}$ is the overall mean. The value of $\eta$ ranges from $[0, 1]$, with values closer to 1 indicating that the categorical variable provides more information about the continuous variable.

\subsection{Categorical-Categorical: Cramér's V}
For two categorical variables with a contingency table of dimensions $r \times k$, Cramér's V is based on the $\chi^2$ statistic. To ensure robustness, we employ the bias correction proposed by \citet{bergsma2013bias}.

First, the bias-corrected $\phi^2$ is calculated as:
\begin{equation}
    \phi^2_{\text{corr}} = \max\left(0, \frac{\chi^2}{n} - \frac{(k-1)(r-1)}{n-1}\right)
\end{equation}
where $n$ is the total number of samples. The corrected dimensions are:
\begin{equation}
    r_{\text{corr}} = r - \frac{(r-1)^2}{n-1}, \quad k_{\text{corr}} = k - \frac{(k-1)^2}{n-1}
\end{equation}
The final Cramér's V score is then given by:
\begin{equation}
    V_{\text{corr}} = \sqrt{\frac{\phi^2_{\text{corr}}}{\min(k_{\text{corr}}-1, r_{\text{corr}}-1)}}
\end{equation}
The value ranges from $[0, 1]$, representing the strength of association between the two categorical variables.

\section{Complete Algorithm}
\label{sec:algorithm}
\label{app:algorithm}

Algorithm~\ref{alg:structsynth} provides the complete pseudocode for the \textsf{StructSynth} framework, integrating both the Evidence-Grounded Graph Induction and the Graph-Planned Conditional Synthesis stages.

\begin{algorithm*}
\caption{\textsf{StructSynth}: Structure-Aware Tabular Data Synthesis}
\label{alg:structsynth}
\begin{algorithmic}[1]
\State \textbf{Input:} Training data $\mathcal{D}_{\mathtt{train}} = (X_{\mathtt{train}}, A)$, number of synthetic samples $s$
\State \textbf{Output:} Synthetic dataset $\mathcal{D}_{\mathtt{synth}} = (X_{\mathtt{synth}}, A)$

\vspace{0.2cm}
\Statex \textbf{Stage 1: Evidence-Grounded Graph Induction}
\Statex \textit{1. Graph Initialization}
\State Select a few-shot subset $\mathcal{D}_{\mathtt{few\_shot}} \subseteq \mathcal{D}_{\mathtt{train}}$
\State $V_{\mathtt{source}} \gets f_{\mathtt{LLM}}(\pi_{\mathtt{source}}(\mathcal{D}_{\mathtt{few\_shot}}))$
\State $G \gets (V_{\mathtt{source}}, \emptyset)$
\State $Q \gets V_{\mathtt{source}}$ \Comment{Initialize queue for BFS}
\State $V_{\mathtt{visited}} \gets \emptyset$

\Statex \textit{2. Iterative Expansion, Resolution, and Continuation (BFS)}
\While{$Q \neq \emptyset$}
    \Statex \hspace{\algorithmicindent} \textit{a. Expansion and Reasoned Link Generation}
    \State $A_i \gets \mathtt{Dequeue}(Q)$
    \State $V_{\mathtt{visited}} \gets V_{\mathtt{visited}} \cup \{A_i\}$
    \State Compute association scores $\mathcal{S}(A_i)$ between $A_i$ and other attributes
    \State $P_i \gets f_{\mathtt{LLM}}(\pi_{\mathtt{generate}}(A_i, G, \mathcal{S}(A_i), \mathcal{D}_{\mathtt{few\_shot}}))$
    \State $E_{\mathtt{prop},t} \gets \{(A_i \to A_j, r_{ij}) \mid (A_j, r_{ij}) \in P_i\}$
    \State $V_{\mathtt{new},t} \gets \{A_j \mid (A_j, \_) \in P_i \land A_j \notin V\}$
    
    \Statex \hspace{\algorithmicindent} \textit{b. LLM-based Cycle Resolution}
    \State $E_{\mathtt{candidate}} \gets E \cup E_{\mathtt{prop},t}$
    \State $\Psi \gets \mathtt{DetectCycles}(V \cup V_{\mathtt{new},t}, E_{\mathtt{candidate}})$
    \State $E_{\mathtt{pruned}} \gets \emptyset$
    \For{each cycle $\psi \in \Psi$}
        \State $e_{\mathtt{to\_remove}} \gets f_{\mathtt{LLM}}(\pi_{\mathtt{resolve}}(\psi))$
        \State $E_{\mathtt{pruned}} \gets E_{\mathtt{pruned}} \cup \{e_{\mathtt{to\_remove}}\}$
    \EndFor
    
    \Statex \hspace{\algorithmicindent} \textit{c. Continuation of BFS}
    \State $G \gets (V \cup V_{\mathtt{new},t}, E_{\mathtt{candidate}} \setminus E_{\mathtt{pruned}})$ \Comment{Update graph}
    \State Enqueue unvisited nodes from $V_{\mathtt{new},t}$ into $Q$
\EndWhile
\State \textbf{Return} $G = (V, E)$

\vspace{-0.1cm}

\Statex \textbf{Stage 2: Graph-Planned Conditional Synthesis}
\State Partition nodes $V$ into topological layers $\mathcal{L} = (L_1, \dots, L_m)$
\State $X_{\mathtt{synth}} \gets \emptyset$
\For{$j \gets 1$ to $s$}
    \Statex \hspace{\algorithmicindent} \textit{a. Generating Graph-Based Values}
    \State $\tilde{x}_{j,V} \gets \{\}$
    \For{$i \gets 1$ to $m$}
        \State $L_i \gets \mathcal{L}[i]$
        \State $\tilde{x}_{j,L_i} \gets f_{\mathtt{LLM}}\Big(\pi_{\mathtt{data\_gen}}(\tilde{x}_{j,<i}, G[L_i \cup \Gamma^-(L_i)], L_i, \mathcal{D}_{\mathtt{few\_shot}}^i)\Big)$
        \State $\tilde{x}_{j,V} \gets \tilde{x}_{j,V} \cup \tilde{x}_{j,L_i}$
    \EndFor
    
    \Statex \hspace{\algorithmicindent} \textit{b. Generating Independent Values}
    \State $A_{\mathtt{iso}} \gets A \setminus V$
    \State $\tilde{x}_{j, A_{\mathtt{iso}}} \gets f_{\mathtt{LLM}}(\pi_{\mathtt{iso}}(\tilde{x}_{j,V}, A_{\mathtt{iso}}, \mathcal{D}_{\mathtt{few\_shot}}^{\mathtt{iso}}))$
    
    \Statex \hspace{\algorithmicindent} \textit{c. Final Assembly}
    \State $\tilde{x}_j \gets \mathtt{concat}(\tilde{x}_{j,V}, \tilde{x}_{j,A_{\mathtt{iso}}})$
    \State $X_{\mathtt{synth}} \gets X_{\mathtt{synth}} \cup \{\tilde{x}_j\}$
\EndFor
\State \textbf{Return} $\mathcal{D}_{\mathtt{synth}} = (X_{\mathtt{synth}}, A)$
\end{algorithmic}
\end{algorithm*}

\section{Detailed Experimental Setup}
\label{app:detailed_experimental_setup}

\begin{table*}[t]
\centering
\resizebox{\textwidth}{!}{%
\small
\begin{tabular}{lcccccc}
\toprule
\textbf{Dataset} & \textbf{\# Num Feat.} & \textbf{\# Cat Feat.} & \textbf{\# Classes} & \textbf{Domain} & \textbf{Task} & \textbf{Post-LLM Knowledge Cutoff} \\
\midrule
Adult~\cite{adult_2} & 4 & 11 & 2 & Social/Census & Binary & \textcolor{darksalmon}{\XSolidBrush} \\
Anxiety~\cite{kaggleSocialAnxiety} & 12 & 7 & 3 & Medical/Health & Multi-Class & \textcolor{green(pigment)}{\Checkmark} \\
Compas~\cite{propublicaMachineBias} & 5 & 3 & 2 & Crime & Binary & \textcolor{darksalmon}{\XSolidBrush} \\
Salary~\cite{kaggleGlobalMarket} & 3 & 15 & - & Economy & Regression & \textcolor{green(pigment)}{\Checkmark} \\
Obesity~\cite{obesity} & 7 & 8 & - & Medical/Health & Regression & \textcolor{darksalmon}{\XSolidBrush} \\
Churn~\cite{churn} & 8 & 4 & 2 & Business & Binary & \textcolor{darksalmon}{\XSolidBrush} \\
\bottomrule
\end{tabular}}%
\caption{Summary of Dataset Characteristics.}
\label{tab:dataset_summary}
\end{table*}

\subsection{Datasets}
\label{app:datasets}
The datasets utilized in this study are summarized in Table~\ref{tab:dataset_summary}. They differ in tasks, number of features, and domain. We randomly split each dataset into training and test sets using an 8:2 ratio, yielding $\mathcal{D}_\mathtt{train\text{-}full}$ and $\mathcal{D}_\mathtt{test}$. To simulate data-scarce settings, we further subsample $n$ instances from $\mathcal{D}_\mathtt{train\text{-}full}$ to form the final training set, $\mathcal{D}_\mathtt{train}$. This procedure is repeated ten times with different random seeds (42--51) to ensure robustness.

\subsection{Baselines}
\label{app:baselines}
For baselines available in SynthCity~\cite{qian2023synthcity}, we use the provided implementations with their default hyperparameters. We evaluate GraDe using the authors' official implementation, including HyFD-based functional-dependency extraction, dynamic sparse attention, and the FD-alignment objective. We use the official CLLM implementation with the same LLM settings as \textsf{StructSynth}. For PAFT, we use the official implementation with its DistilGPT2 fine-tuning setup; its outputs are evaluated under the same 100-shot XGBoost protocol. Table~\ref{tab:hyperparameters} reports the shared evaluator settings and the LLM configuration used for CLLM and \textsf{StructSynth}.

\subsection{Implementation and Hyperparameters}
\label{app:hyperparameter}
The detailed hyperparameter configurations for the LLMs in \textsf{StructSynth}, the downstream evaluation models, and the Peter-Clark algorithm are provided in Table~\ref{tab:hyperparameters}. For the software implementation, we utilize the Langchain library to manage LLM inference and the Causal-Learn library~\cite{zheng2024causal} for the PC algorithm.

\begin{table}[t]
\centering
\small
\setlength{\tabcolsep}{4pt}
\begin{tabular}{ll}
    \toprule
    \textbf{Group} & \textbf{Hyperparameter Settings} \\
    \midrule
    \multirow{5}{*}{LLMs} & frequency\_penalty = 0 \\
    & presence\_penalty = 0 \\
    & top\_p = 0.95 \\
    & temperature = 0.9 \\
    & max\_tokens = 8000 \\
    \midrule
    \multirow{10}{*}{XGBoost} & booster = gbtree \\
    & n\_estimators = 100 \\
    & learning\_rate = 0.1 \\
    & max\_depth = 6 \\
    & min\_child\_weight = 1 \\
    & gamma = 0 \\
    & subsample = 1 \\
    & colsample\_bytree = 1 \\
    & reg\_alpha = 0 \\
    & reg\_lambda = 1 \\
    \midrule
    \multirow{2}{*}{Random Forest} & n\_estimators = 200 \\
    & n\_jobs = $-1$ \\
    \midrule
    \multirow{2}{*}{Logistic Reg.} & max\_iter = 1000 \\
    & solver = `lbfgs' \\
    \midrule
    Ridge & alpha = 1.0 \\
    \midrule
    \multirow{2}{*}{MLP} & hidden\_layer\_sizes = (100,) \\
    & max\_iter = 500 \\
    \midrule
    \multirow{3}{*}{PC algorithm} & alpha = 0.05 \\
    & indep\_test = `kci' \\
    & stable = True \\
    \bottomrule
\end{tabular}
\caption{Hyperparameter Settings for Our Experiments.}
\label{tab:hyperparameters}
\end{table}

\subsection{Evaluation Metrics}
\label{app:metrics}

\subsubsection{Downstream Model Performance}
The Downstream Model Performance metric serves as the primary measure of the synthetic data's practical utility. It evaluates whether the synthetic data, when used to augment the original training set, can improve the performance of a machine learning model on a real, unseen test set. For each generative method, we create an augmented training set, $\mathcal{D}_{\text{aug}}$, by combining the original limited training data, $\mathcal{D}_{\text{train}}$, with the generated synthetic data, $\mathcal{D}_{\text{synth}}$:
\begin{equation}
    \mathcal{D}_{\text{aug}} = \mathcal{D}_{\text{train}} \cup \mathcal{D}_{\text{synth}}
\end{equation}
A downstream machine learning model (in our experiments, an XGBoost model) is then trained exclusively on this augmented dataset, $\mathcal{D}_{\text{aug}}$, and evaluated on the held-out real test set, $\mathcal{D}_{\text{test}}$. For binary and multi-class classification tasks, we report the Area Under the Receiver Operating Characteristic Curve (AUC). For regression tasks, we report the R-squared ($R^2$) score.

\subsubsection{Privacy Risk Assessment Metric}
\label{app:privacy_metric}
The Privacy Risk Assessment metric evaluates the extent to which a generative model overfits or memorizes the training data, $\mathcal{D}_{\text{train}}$. The complete set of real data instances is defined as $\mathcal{D}_{\text{real}} = \mathcal{D}_{\text{train}} \cup \mathcal{D}_{\text{test}}$. To ensure an unbiased comparison where the ideal probability is exactly 0.5, we first downsample the larger of the two sets ($\mathcal{D}_{\text{train}}$ or $\mathcal{D}_{\text{test}}$) to match the size of the smaller one.

For any two data points, $x_i$ and $x_j$, from the mixed-type dataset, we define a distance function $d(x_i, x_j)$ as the sum of the L1 distance for numerical features and the Hamming distance for categorical features:
\begin{equation}
\begin{split}
    d(x_i, x_j) ={} & \sum_{k \in \text{numerical}} |x_{i,k} - x_{j,k}| \\
    & + \sum_{l \in \text{categorical}} \mathbb{I}(x_{i,l} \neq x_{j,l})
\end{split}
\end{equation}
For each synthetic sample $\tilde{x} \in \mathcal{D}_{\text{synth}}$, we identify its nearest neighbor ($NN$) from the entire set of real data instances, $\mathcal{D}_{\text{real}}$:
\begin{equation}
    NN(\tilde{x}) = \underset{x \in \mathcal{D}_{\text{real}}}{\arg\min} \, d(\tilde{x}, x)
\end{equation}
The Privacy Risk score is then calculated as the fraction of synthetic samples whose nearest neighbor is an instance from the training set $\mathcal{D}_{\text{train}}$:
\begin{equation}
    \text{PrivacyRisk} = \frac{1}{|\mathcal{D}_{\text{synth}}|} \sum_{\tilde{x} \in \mathcal{D}_{\text{synth}}} \mathbb{I}(NN(\tilde{x}) \in \mathcal{D}_{\text{train}})
\end{equation}
A score approaching 0.5 suggests that the generative model has not memorized the training data.

\subsubsection{Statistical Fidelity Metric}
\label{app:fidelity_metric}
The Statistical Fidelity metric quantifies how well the synthetic data, $\mathcal{D}_{\text{synth}}$, preserves the pairwise statistical relationships present in the real data, $\mathcal{D}_{\text{real}}$. For each unique pair of columns $(C_i, C_j)$ in the dataset, a difference score, $\Delta(C_i, C_j)$, is calculated. The calculation method varies for the three possible data type combinations:

\begin{itemize}[leftmargin=*]
    \item {\textbf{Numerical-Numerical Pairs:} For two numerical columns, the difference is the absolute difference between their Pearson correlation coefficients ($\rho$) in the real and synthetic datasets:
\begin{equation}
\begin{split}
    \Delta(C_i, C_j) = \big| \, & \rho(C_{i, \text{real}}, C_{j, \text{real}}) \\
    -\ & \rho(C_{i, \text{synth}}, C_{j, \text{synth}}) \big|
\end{split}
\end{equation}
}
    \item {\textbf{Categorical-Categorical Pairs:} For two categorical columns, the difference is measured as the Total Variation Distance (TVD) of their joint probability distributions:
\begin{align}
    \Delta(C_i, C_j) = \frac{1}{2} \sum_{u,v} \big| 
    & P_{\text{real}}(C_i=u, C_j=v) \notag \\
    -\ & P_{\text{synth}}(C_i=u, C_j=v) \big|
\end{align}
where $u$ and $v$ iterate over all possible categories for columns $C_i$ and $C_j$, respectively.}
    \item {\textbf{Numerical-Categorical Pairs:} For a mixed pair of a numerical column $C_i$ and a categorical column $C_j$, the numerical column is first discretized into a categorical representation, $C'_i$, by binning its values into $q=10$ quantiles. The difference is then calculated using the same TVD method as for categorical-categorical pairs:
\begin{equation}
    \Delta(C_i, C_j) = \Delta(C'_i, C_j)
\end{equation}}
\end{itemize}

The overall Statistical Fidelity score is the mean of these difference scores computed over all unique pairs of columns in the dataset:
\begin{equation}
    \text{StatisticalFidelity} = \frac{1}{\binom{K}{2}} \sum_{1 \le i < j \le K} \Delta(C_i, C_j)
\end{equation}
where $K$ is the total number of columns.

\section{Qualitative Graph Analysis}
\label{app:qualitative_graph_analysis}

\subsection{Structural Fidelity on Adult Dataset}
\label{app:adult_graph}
\begin{figure*}[t!]
    \centering
    \begin{subfigure}[b]{0.37\textwidth}
        \centering
        \includegraphics[width=\linewidth]{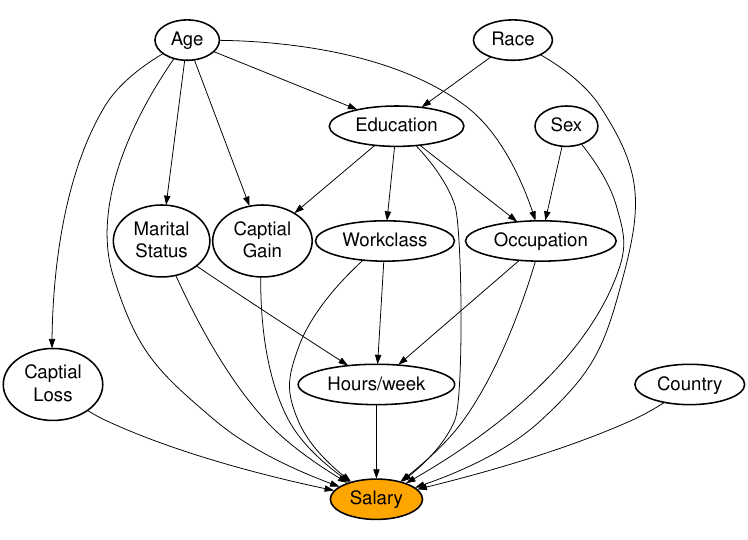}
        \caption{Reference Graph}
        \label{fig:structure_groundtruth}
    \end{subfigure}%
    \hfill 
    \begin{subfigure}[b]{0.3\textwidth}
        \centering
        \includegraphics[width=\linewidth]{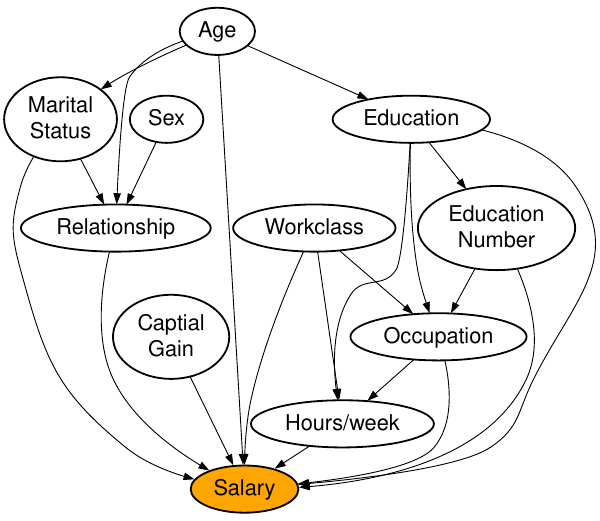}
        \caption{LLM-Discovered}
        \label{fig:structure_llmcd}
    \end{subfigure}%
    \hfill 
    \begin{subfigure}[b]{0.32\textwidth}
        \centering
        \includegraphics[width=\linewidth]{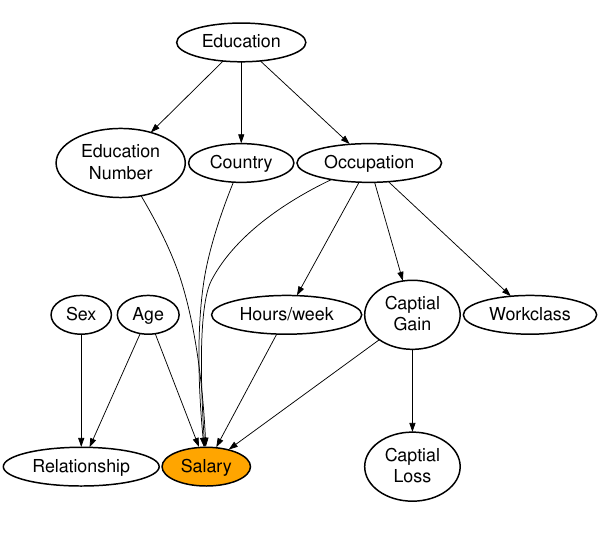}
        \caption{Re-discovered from Synthetic Data}
        \label{fig:structure_structsynth}
    \end{subfigure}
    \caption{Visual Analysis of Structural Fidelity on the Adult Dataset with \textit{Salary} as the Label Column.}
    \label{fig:structure}
\end{figure*}

To provide a qualitative illustration of our method's ability to preserve the core dependencies within the data in the absence of a ground-truth graph, we present a case study on the Adult dataset in Figure~\ref{fig:structure}. This visualization compares three distinct graphs: (a) a reference graph, established by running the PC algorithm on the complete dataset and refining the output with domain expertise; (b) the dependency graph discovered by \textsf{StructSynth}'s learning stage from $\mathcal{D}_\mathtt{train}$; and (c) a graph re-discovered from $\mathcal{D}_\mathtt{synth}$ generated by \textsf{StructSynth} using the PC algorithm.

A visual comparison reveals a strong alignment between our LLM-discovered graph (Figure~\ref{fig:structure_llmcd}) and the reference graph (Figure~\ref{fig:structure_groundtruth}), particularly around the key \textit{Salary} variable. More importantly, the re-discovered graph (Figure~\ref{fig:structure_structsynth}) recovers the same core structure: a standard discovery algorithm applied to our synthetic data returns essentially the reference graph, indicating that \textsf{StructSynth} preserves the dependency structure rather than only the marginals.

\subsection{Structure Discovery Comparisons on Asia Dataset}
\label{app:structure_examples}
\begin{figure*}[t!]
\centering
\includegraphics[width=\textwidth]{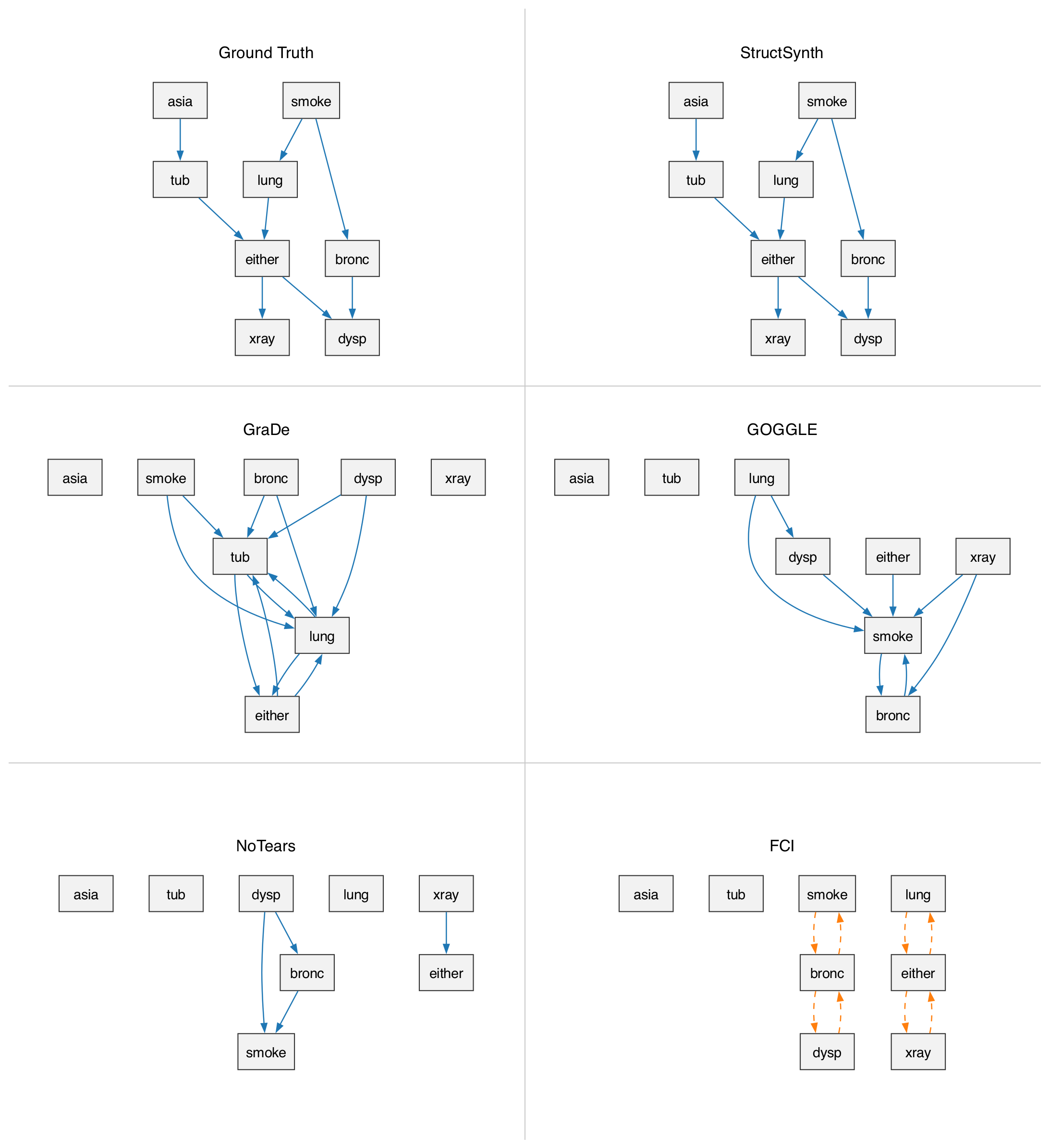}
\caption{Comparison of discovered dependency structures on the Asia dataset ($n=100$). \textbf{Ground Truth}: The reference DAG from the bnlearn repository. \textbf{StructSynth}: Our method closely recovers the ground truth. \textbf{GraDe, GOGGLE, NoTears, FCI}: Baseline methods exhibit varying degrees of structural errors, including missing edges, reversed edges, and spurious connections.}
\label{fig:structure_examples}
\end{figure*}

To provide a concrete illustration of the structural discovery capabilities discussed above, Figure~\ref{fig:structure_examples} presents a visual comparison of the dependency graphs learned by \textsf{StructSynth} and four representative baselines on the Asia dataset ($n=100$). Although the Asia benchmark is defined as a causal Bayesian network, \textsf{StructSynth} uses the ground-truth DAG only as a reference directed dependency structure for SHD evaluation. The \textbf{Ground Truth} graph represents the reference dependency structure of the dataset, where, for example, \textit{smoke} is linked to \textit{lung} (cancer) and \textit{bronc} (bronchitis), which in turn are linked to \textit{dysp} (dyspnoea).

\textsf{StructSynth} successfully recovers the majority of these key dependencies, including the V-structure at \textit{dysp} and the dependency chain from \textit{smoke} to \textit{lung}. The slight deviations are minimal and do not disrupt the global topological ordering, ensuring that the subsequent generation process follows a valid dependency path.

In contrast, baseline methods exhibit significant structural distortions:
\begin{itemize}[leftmargin=*]
    \item \textbf{GraDe}'s extracted FD graph is denser than the reference DAG in this $n=100$ example and contains spurious edges, such as a direct link between \textit{asia} and \textit{dysp}.
    \item \textbf{GOGGLE}, while capturing some dependencies, suffers from edge reversals (e.g., \textit{lung} $\rightarrow$ \textit{smoke}) and misses critical connections (e.g., \textit{bronc} is isolated from \textit{dysp}). This reflects the instability of end-to-end differentiable structure learning in low-data regimes.
    \item \textbf{NoTears} yields a sparse and fragmented graph, failing to identify the relationship between \textit{smoke} and \textit{lung}, which are central to the dataset's dependency structure.
    \item \textbf{FCI}, a constraint-based method, correctly identifies some skeletons but leaves many edges undirected (represented by bi-directed or unoriented edges), failing to provide a clear generative order.
\end{itemize}

These visual examples corroborate our quantitative findings in Figure~\ref{fig:shd}, confirming that \textsf{StructSynth}'s LLM-guided approach offers superior robustness in discovering interpretable and accurate dependency structures compared to both traditional and deep learning-based alternatives.

\section{Additional Results on Post-Cutoff Datasets}
\label{app:post_cutoff_robustness}

\paragraph{Sensitivity to Training Sample Size.}
Complementing the Adult analysis in Figure~\ref{fig:influence_n}, Table~\ref{tab:post_cutoff_vary_n} reports downstream utility on Anxiety and Salary across $n\in\{20,50,100,200\}$.

\begin{table}[t]
\centering
\small
\setlength{\tabcolsep}{3.5pt}
\resizebox{\columnwidth}{!}{%
\begin{tabular}{llcccc}
\toprule
\textbf{Dataset} & \textbf{Method} & $n=20$ & $n=50$ & $n=100$ & $n=200$ \\
\midrule
\multirow{2}{*}{Anxiety (AUC)}
  & CLLM        & 77.52          & 81.87          & 85.17          & 85.44 \\
  & StructSynth & \textbf{80.83} & \textbf{83.50} & \textbf{86.45} & \textbf{87.23} \\
\midrule
\multirow{2}{*}{Salary ($R^2$)}
  & CLLM        & 48.16          & 52.34          & 54.53          & 61.39 \\
  & StructSynth & \textbf{50.56} & \textbf{54.90} & \textbf{55.98} & \textbf{63.54} \\
\bottomrule
\end{tabular}
}
\caption{Downstream utility across training-set sizes on the post-cutoff datasets. Both methods generate 1,000 synthetic rows per run. Anxiety AUC and Salary $R^2$ are reported on a 0--100 scale. The higher reported score in each comparison is in \textbf{bold}.}
\label{tab:post_cutoff_vary_n}
\end{table}

\paragraph{LLM Backbone.}
Table~\ref{tab:qwen25_post_cutoff} reports an additional evaluation with Qwen2.5-32B at $n=100$ on the two post-cutoff datasets.

\begin{table}[t]
\centering
\small
\setlength{\tabcolsep}{4pt}
\begin{tabular}{llcc}
\toprule
\textbf{Dataset} & \textbf{Metric} & \textbf{CLLM} & \textbf{StructSynth} \\
\midrule
Anxiety & AUC   & $0.843 \pm 0.012$ & $\mathbf{0.856} \pm 0.014$ \\
Salary  & $R^2$ & $0.518 \pm 0.037$ & $\mathbf{0.532} \pm 0.036$ \\
\bottomrule
\end{tabular}
\caption{Downstream utility on the post-cutoff datasets with Qwen2.5-32B at $n=100$ and 1,000 synthetic rows per run. Results are reported on a 0--1 scale as mean $\pm$ standard deviation across runs; the higher mean for each dataset is in \textbf{bold}.}
\label{tab:qwen25_post_cutoff}
\end{table}

\section{Efficiency Analysis}
\label{sec:token_usage}

A practical concern is the additional cost introduced by structure discovery. Table~\ref{tab:token_usage} reports the token consumption of \textsc{CLLM} and \textsf{StructSynth} under the \textbf{100-shot} setting, averaged over six datasets. Overall, introducing the graph-based structure modeling in \textsf{StructSynth} does \emph{not} lead to a prohibitive cost increase: the average total usage rises from \textbf{219.8K} tokens (\textsc{CLLM} total) to \textbf{287.9K} tokens (\textsf{StructSynth} combined total), corresponding to an absolute overhead of \textbf{+68.1K} tokens (about \textbf{+31.0\%}). This overhead is relatively stable across datasets (roughly \textbf{+19\%} to \textbf{+42\%} in total), suggesting that the additional structure reasoning introduces a consistent but moderate computational budget.

A key observation is that the overhead is \textbf{highly skewed toward the input side}. Concretely, \textsf{StructSynth} increases average \emph{input} tokens by \textbf{+65.1K} (\textbf{+57.1\%}), while the \emph{output} tokens only increase by \textbf{+2.9K} (\textbf{+2.7\%}). This pattern is expected: \textsf{StructSynth} needs additional prompt context to (i) elicit and refine the dependency graph and (ii) condition the synthesis stage on the discovered structure, whereas the corresponding model responses (e.g., edge lists / compact graph representations) are short and the final table synthesis outputs remain comparable in length to \textsc{CLLM}.

Importantly, this ``input-dominant'' overhead implies that the \textbf{real monetary overhead is often smaller than what the raw total-token increase might suggest}, because in many pricing schemes output tokens are priced higher than input tokens. Taken together, the token analysis indicates that \textsf{StructSynth} achieves structure-aware generation with a \textbf{modest and practically manageable} cost increase, largely attributable to additional prompting rather than longer model generations.

\begin{table*}[t]
  \centering
  \resizebox{\textwidth}{!}{%
  \small
  \renewcommand{\arraystretch}{1.12}
  \begin{tabular}{@{}ll *{7}{S[table-format=+3.1]}@{}}
    \toprule
    \textbf{Method} & \textbf{Component} & {\textbf{Adult}} & {\textbf{Anxiety}} & {\textbf{Churn}} & {\textbf{Compas}} & {\textbf{Obesity}} & {\textbf{Salary}} & {\textbf{Avg.}} \\
    \midrule
    \multirow{3}{*}{\textsc{CLLM}}
      & Input  &  40.6 & 142.2 &  38.3 &  26.5 &  70.7 & 366.1 & 114.1 \\
      & Output & 105.2 & 175.2 & 102.6 &  79.5 &  13.5 & 158.6 & 105.8 \\
      & {\bfseries Total}
               & {\bfseries 145.8} & {\bfseries 317.4} & {\bfseries 140.9} & {\bfseries 106.0}
               & {\bfseries  84.2} & {\bfseries 524.7} & {\bfseries 219.8} \\
    \midrule
    \multirow{7}{*}{\textsf{StructSynth}}
      & Graph Input  & 38.1 &  89.0 & 41.4 & 17.1 & 43.5 & 111.1 & 56.7 \\
      & Graph Output &  1.5 &   3.1 &  2.2 &  1.0 &  1.7 &   3.1 &  2.1 \\
      \rowcolor{SubtotalGray}
      & Graph total  & 39.6 &  92.1 & 43.6 & 18.1 & 45.3 & 114.3 & 58.8 \\
      \cmidrule(lr){2-9}
      & Table Input  &  52.7 & 168.3 &  46.0 &  32.3 &  61.1 & 374.3 & 122.5 \\
      & Table Output & 114.3 & 177.4 & 102.7 &  75.9 &  12.0 & 157.1 & 106.6 \\
      \rowcolor{SubtotalGray}
      & Table total  & 167.0 & 345.7 & 148.7 & 108.2 &  73.2 & 531.4 & 229.0 \\
      \cmidrule(lr){2-9}
      & {\bfseries Combined total}
                   & {\bfseries 206.6} & {\bfseries 437.8} & {\bfseries 192.3} & {\bfseries 126.3}
                   & {\bfseries 118.4} & {\bfseries 645.7} & {\bfseries 287.9} \\
    \midrule
    \multicolumn{9}{@{}l}{\textbf{Token overhead of \textsf{StructSynth} vs. \textsc{CLLM} (Combined $-$ Total)}} \\
      & $\Delta$ Input (K)  &  50.2 & 115.1 &  49.1 &  22.9 &  33.9 & 119.3 &  65.1 \\
      & $\Delta$ Output (K) &  10.6 &   5.3 &   2.3 &  -2.6 &   0.2 &   1.6 &   2.9 \\
      & $\Delta$ Total (K)  &  60.8 & 120.4 &  51.4 &  20.3 &  34.2 & 121.0 &  68.1 \\
      \cmidrule(lr){2-9}
      & Relative Input (\%)  & 123.6 &  80.9 & 128.2 &  86.4 &  48.0 &  32.6 &  57.1 \\
      & Relative Output (\%) &  10.1 &   3.0 &   2.2 &  -3.3 &   1.5 &   1.0 &   2.7 \\
      & Relative Total (\%)  &  41.7 &  37.9 &  36.5 &  19.2 &  40.6 &  23.1 &  31.0 \\
    \bottomrule
  \end{tabular}}%
  \caption{Token usage under 100-shot ($\times 10^3$ tokens). We report input/output tokens for \textsc{CLLM} and \textsf{StructSynth}; overhead compares \textsf{StructSynth} combined totals (graph+table) against \textsc{CLLM} totals.}
  \label{tab:token_usage}
\end{table*}

\section{Extended Discussion}
\label{app:extended_discussion}

In this section, we discuss the broader implications of our results, analyze key design decisions, and contextualize \textsf{StructSynth} within the landscape of structure-aware generative methods.

\subsection{Structure-Blind vs.\ Structure-Aware Generation}
\label{sec:structure_discussion}
A central question raised by our work is whether explicit structural guidance provides a \emph{qualitative}---rather than merely quantitative---advantage over structure-blind LLM generation. To address this, we compare \textsf{StructSynth} against its closest structure-blind counterpart, CLLM, which uses the same LLM with identical hyperparameters but generates data without any structural generation plan.

Our ablation study (Table~\ref{tab:ablation_study}) provides direct evidence for three distinct dimensions of improvement. First, regarding \textbf{structural modeling}: CLLM processes features in an arbitrary serialized order, implicitly inducing dependencies from the input sequence. In contrast, \textsf{StructSynth} encodes dependencies as an explicit DAG, ensuring that relationships are \emph{discovered} rather than incidentally induced. Second, regarding \textbf{generation strategy}: CLLM generates all features in a single pass without constraints, whereas \textsf{StructSynth} follows a topological ordering so that each feature is conditioned on its discovered parent nodes. Removing this ordering alone (\textit{No Topological Order} variant) degrades AUC by 1.1 points, while removing the graph entirely (\textit{No Structure} variant, equivalent to CLLM) causes a 1.6-point drop. Third, regarding \textbf{regularization}: the DAG serves as a structural regularizer that constrains the generation space, preventing the LLM from overfitting to spurious patterns in the few available training samples. This regularization effect is evidenced by \textsf{StructSynth}'s superior privacy preservation (avg.\ rank 1.50 vs.\ CLLM's 3.50), indicating reduced memorization of training records.

These results indicate that the advantage of \textsf{StructSynth} comes from how dependencies are modeled and enforced during generation, rather than from peripheral additions to an LLM pipeline. We note that the largest single ablation effect is the loss of the LLM generator itself ($-4.4$ pts), so the generation plan should be read as organizing a strong generator rather than as the dominant source of performance.

\subsection{Understanding the Fidelity-Privacy Trade-off}
\label{sec:fidelity_discussion}
A nuanced finding in our evaluation is that \textsf{StructSynth} does not achieve the highest Statistical Fidelity score (avg.\ rank 9.50), despite its strong downstream utility (avg.\ rank 1.00) and privacy preservation (avg.\ rank 1.50). We argue that this apparent discrepancy is both expected and desirable, reflecting the inherent tension between fidelity and privacy.

To understand this trade-off, consider GReaT, which achieves strong Statistical Fidelity (avg.\ rank 4.17). However, its Privacy Risk score of 90.15\% reveals severe memorization---the model essentially reproduces training records rather than generating novel samples. This high fidelity is therefore an artifact of overfitting rather than genuine distributional learning. At the other extreme, the Bayesian Sampler in our ablation achieves the best raw fidelity score among all variants (48.86) but the lowest AUC (81.17), demonstrating that \textbf{high pairwise fidelity does not guarantee downstream utility}.

This discrepancy can be further understood by examining what Statistical Fidelity actually measures. The metric quantifies the preservation of \emph{pairwise} correlations across all feature pairs. In contrast, \textsf{StructSynth}'s DAG encodes \emph{conditional} dependencies---the probabilistic relationships between parent and child nodes. These conditional dependencies are what a downstream model relies on. \textsf{StructSynth} optimizes for preserving these conditional structures rather than pairwise statistics, which explains its superior utility despite moderate pairwise fidelity scores.

In summary, \textsf{StructSynth}'s generation plan acts as a \emph{regularizer}: it constrains the generation process to follow discovered conditional dependencies, which simultaneously prevents memorization (ensuring privacy) and preserves the dependency structure that matters most for downstream tasks (ensuring utility), even at the cost of slightly lower pairwise correlation matching.

\subsection{When Does Structural Guidance Help Most?}
\label{sec:when_helps}
Our experiments on varying training sample sizes (Figure~\ref{fig:influence_n}) reveal that \textsf{StructSynth}'s advantage is most pronounced in the low-data regime ($n \leq 50$), where it maintains high AUC while baselines deteriorate significantly. As $n$ increases toward 200, the performance gap narrows, though \textsf{StructSynth} remains competitive.

This pattern has an intuitive explanation. When data is scarce, statistical methods for structure discovery (e.g., PC algorithm, NoTears) suffer from unstable association estimates, leading to unreliable graphs. In this regime, \textsf{StructSynth}'s LLM-guided discovery provides a powerful complement: the model's semantic prior knowledge acts as a \emph{structural regularizer}, stabilizing the discovery process by filtering out statistically spurious correlations that would mislead purely data-driven approaches. As more data becomes available, statistical methods become increasingly reliable on their own, diminishing the relative advantage of LLM-guided reasoning.

This analysis also reveals \textsf{StructSynth}'s natural boundary conditions. The method is best suited for scenarios where: (1) the number of training samples is severely limited ($n \leq 100$); (2) features have semantically meaningful names and relationships that an LLM can reason about; and (3) the underlying dependency structure is sparse enough to be well-represented by a DAG. In contrast, for purely numerical datasets with opaque feature names (e.g., anonymized sensor readings), the LLM's semantic prior provides less value, and the framework may offer diminished advantages over purely statistical approaches.

\subsection{Relationship to Causal Inference}
It is important to clearly distinguish \textsf{StructSynth}'s dependency graphs from causal graphs in the traditional sense. As established in Appendix~\ref{sec:why_dag}, our DAG serves as a \emph{generative dependency structure}---its edges encode statistical and probabilistic dependencies used to organize the synthesis process, not claims of ground-truth causality.

This distinction separates \textsf{StructSynth} from classical causal discovery methods such as the PC algorithm~\cite{spirtes2000causation}, FCI~\cite{spirtes2000causation}, or GES~\cite{chickering2002optimal}, which aim to identify the true causal data-generating mechanism under specific assumptions (e.g., faithfulness, sufficiency). Our objective is fundamentally different: we seek a dependency structure that enables \emph{effective conditional generation}, not one that supports causal interventions or counterfactual reasoning.

Nevertheless, our SHD evaluations on benchmark DAGs (Figure~\ref{fig:shd}) show that the structures discovered by \textsf{StructSynth} align closely with the reference dependency structures of those benchmarks. This is not a contradiction: when a dataset is generated from a DAG, a good dependency structure will largely coincide with that DAG's edges. This alignment reflects the structural quality of the induced plan---which is what the generation plan is meant to capture---not the pairwise correlation matching that Statistical Fidelity measures.

\section{Detailed Prompts for StructSynth}
\label{app:prompts}
This section provides a guide to the prompt templates used in the \textsf{StructSynth} framework. A key step in our methodology is the textualization of diverse data structures---including statistical scores, dependency graphs, and tabular data---to serve as inputs for the Large Language Model (LLM). \hyperref[box:textualization_examples]{Box 1} shows concrete examples of this process. 

The \textsf{StructSynth} pipeline employs a sequence of prompts to complete its tasks. First, the \hyperref[prompt:source]{\texttt{$\pi_\mathtt{source}$}} prompt initiates graph discovery by identifying initial root nodes. During the iterative graph construction, the \hyperref[prompt:generation]{\texttt{$\pi_\mathtt{generate}$}} prompt proposes new dependency edges with corresponding rationales. If a cycle is detected, the \hyperref[prompt:resolve]{\texttt{$\pi_\mathtt{resolve}$}} prompt is used to analyze the conflict and remove the weakest link. Finally, for data synthesis, the \hyperref[prompt:data_gen]{\texttt{$\pi_\mathtt{data\_gen}$}} prompt generates values for dependent features, after which the \hyperref[prompt:data_gen_iso]{\texttt{$\pi_\mathtt{iso}$}} prompt generates values for the remaining isolated features.

\begin{tcolorbox}[
    float*=t, width=\textwidth, floatplacement=htbp,
    colback=teal!10!white,
    colframe=teal!60!black,
    title={Examples of Data Textualization for LLM Prompts},
    fonttitle=\bfseries\color{white},
    arc=2mm,
    boxrule=1pt
] 

\small

\textbf{1. Textual Representation of Statistical Evidence (Association Scores):}
\begin{verbatim}
Association scores for 'Education', scaled 0 to 1.
(Levels: >0.8 Very High, >0.6 High, >0.4 Moderate, >0.2 Low, <=0.2 Very Low)

- Occupation: 0.85 (Very High)
- Salary: 0.72 (High)
- Marital-Status: 0.55 (Moderate)
- Hours-per-week: 0.31 (Low)
- Sex: 0.12 (Very Low)
\end{verbatim}

\vspace{\medskipamount}

\textbf{2. Textual Representation of a Dependency Graph:}
\begin{verbatim}
Current dependency graph structure with rationales:
- Age -> Marital-Status
Rationale: A person's age is a primary factor influencing their marital status.

- Sex -> Relationship: 
Rationale: Gender often determines the relationship role within a household.

- Country -> Race: 
Rationale: Native country has a strong statistical correlation with a person's race.

- Race -> Education: 
Rationale: Statistical data often shows a correlation between a person's race 
and their level of educational attainment.

\end{verbatim}

\vspace{\medskipamount}

\textbf{3. Textual Representation of Tabular Data (Markdown):}
\begin{verbatim}
| age | education   | occupation      | sex    | hours-per-week | salary |
|-----|-------------|-----------------|--------|----------------|--------|
| 39  | Bachelors   | Adm-clerical    | Male   | 40             | <=50K  |
| 50  | Bachelors   | Exec-managerial | Male   | 13             | <=50K  |
| 28  | Assoc-acdm  | Prof-specialty  | Female | 45             | <=50K  |
\end{verbatim}
\label{box:textualization_examples}
\end{tcolorbox}

\begin{tcolorbox}[
    float*=t, width=\textwidth, floatplacement=htbp,
    colback=yellow!10!white,
    colframe=yellow!75!black,
    title={Prompt for Source Nodes Initialization of Tabular Data ($\pi_\mathtt{source}$)},
    fonttitle=\bfseries\color{white},
    arc=2mm,
    boxrule=1pt
]
\small

You are an expert data analyst. Based on the following information:

\vspace{\medskipamount}

\textbf{Task Description:}

Identify the initial ``source nodes'' for a dependency graph. A source node is a root attribute that is generated without conditioning on other attributes in this dataset.

\vspace{\medskipamount}

\textbf{All Feature Descriptions:}
\begin{verbatim}
{all_features}
\end{verbatim}

\vspace{\medskipamount}

\textbf{Example Data:}
\begin{verbatim}
{few_shot_data}
\end{verbatim}

\vspace{\medskipamount}

\textbf{Your Task is to Identify the Source Nodes:}
\begin{itemize}[leftmargin=*, itemsep=2pt]
    \item Your only task is to analyze the provided column descriptions and data.
    \item Identify the features that are most likely to be \textbf{source nodes} (i.e., fundamental attributes that are not dependent on other features).
    \item Provide your answer as a simple list of the identified source node names.
\end{itemize}
\label{prompt:source}
\end{tcolorbox}

\begin{tcolorbox}[
    float*=t, width=\textwidth, floatplacement=htbp,
    colback=pink!10!white,
    colframe=pink!75!black,
    title={Prompt for Link Generation in Dependency Discovery ($\pi_\mathtt{generate}$)},
    fonttitle=\bfseries\color{white}, 
    arc=2mm,
    boxrule=1pt
]
\small

You are an expert data analyst building a dependency graph. Based on the following information:

\vspace{\medskipamount}

\textbf{Task Description:}

For the given $\texttt{\{current\_feature}\}$, your task is to propose its direct successors (dependents) 
from the list of available features. For each proposed dependency, you must provide a 
brief, evidence-based rationale.

\vspace{\medskipamount}

\textbf{Current Graph State:}
\begin{verbatim}
{structure_graph}
\end{verbatim}
\vspace{\medskipamount}

\textbf{All Candidate Features:}
\begin{verbatim}
{all_features}
\end{verbatim}
\vspace{\medskipamount}

\textbf{Statistical Evidence (Association Scores):}
\begin{verbatim}
{association_scores}
\end{verbatim}
\vspace{\medskipamount}

\textbf{Example Data:}
\begin{verbatim}
{few_shot_data}
\end{verbatim}
\vspace{\medskipamount}

\textbf{Your Task is to Propose and Justify New Links:}
\begin{itemize}[leftmargin=*, itemsep=2pt]
    \item Your focus is on the feature: $\texttt{\{current\_feature}\}$.
    \item Review the \textbf{Statistical Evidence}, which shows the strength of association between $\texttt{\{current\_feature\}}$ and other candidate features $\texttt{\{all\_features\}}$.
    \item Based on the statistical scores, the example data, and the existing graph, identify which other features are most likely to be \textbf{direct dependents} of $\texttt{\{current\_feature\}}$.
    \item For each dependency you propose, you \textbf{must} provide a concise $\texttt{rationale}$. This rationale should explain why the dependency makes sense (e.g., ``Higher Education is strongly correlated with and typically precedes higher Income'').
    \item Do not propose links that would create an obvious logical contradiction with the existing graph structure.
    \item Format your final output as a list of dependency edges and their corresponding rationales.
\end{itemize}

\label{prompt:generation}
\end{tcolorbox}

\begin{tcolorbox}[
    float*=t, width=\textwidth, floatplacement=htbp,
    colback=green!10!white,
    colframe=green!60!black,
    title={Prompt for Cycle Resolution in Dependency Discovery ($\pi_\mathtt{resolve}$)},
    fonttitle=\bfseries\color{white},
    arc=2mm,
    boxrule=1pt
]
\small

You are a logical reasoning expert tasked with resolving a contradiction in a dependency graph. Based on the following information:

\vspace{\medskipamount}

\textbf{Task Description:}

Analyze the provided dependencies and their rationales within the cycle. Identify the 
single dependency with the weakest or least plausible rationale and recommend it for 
removal to break the cycle.

\vspace{\medskipamount}

\textbf{Conflicting Cycle with Rationales:}
\begin{verbatim}
{cycle_with_rationales}
\end{verbatim}
\vspace{\medskipamount} 

\textbf{Your Task is to Identify the Weakest Link:}
\begin{itemize}[leftmargin=*, itemsep=2pt]
    \item A logical contradiction (a cycle) has been detected in the graph, detailed in \textbf{Conflicting Cycle with Rationales}.
    \item Your task is to carefully evaluate the \texttt{rationale} provided for each edge in this cycle.
    \item Based on your analysis of the rationales and any relevant context from the column descriptions or data, identify the \textbf{single weakest link}.
    \item The weakest link is the dependency whose rationale is the least plausible, least supported, or most likely to be spurious.
    \item Provide your final answer as the single dependency edge that should be removed to resolve the contradiction.
\end{itemize}

\label{prompt:resolve}
\end{tcolorbox}

\begin{tcolorbox}[
    float*=t, width=\textwidth, floatplacement=htbp,
    colback=blue!5!white,
    colframe=blue!75!black,
    title={Prompt for Graph-Planned Conditional Synthesis ($\pi_\mathtt{data\_gen}$)},
    fonttitle=\bfseries\color{white},
    arc=2mm,
    boxrule=1pt
]
\small 

You are a helpful AI assistant that generates realistic tabular data based on structural dependencies.

\vspace{\medskipamount}

\textbf{Task Description:}

Generate a realistic synthetic table containing exactly the columns listed in 
$\texttt{\{features\_to\_generate\}}$. This generation must be conditioned on the provided values 
of their parent features.

\vspace{\medskipamount}

\textbf{Conditioning Data (Current Results):}
\begin{verbatim}
{current_result}
\end{verbatim}
\vspace{\medskipamount}

\textbf{Relevant Dependency Structure:}
\begin{verbatim}
{related_structure_graph}
\end{verbatim}
\vspace{\medskipamount}

\textbf{Example Data:}
\begin{verbatim}
{related_few_shot_data}
\end{verbatim}
\vspace{\medskipamount}

\textbf{Your Task is to Generate the Synthetic Data:}
\begin{itemize}[leftmargin=*, itemsep=2pt]
    \item Your primary task is to generate realistic data for the features listed in $\texttt{\{features\_to\_generate\}}$.
    \item This generation must be \textbf{conditioned} on the data provided in \textbf{Conditioning Data (Current Results)}. These are the values of the parent nodes that influence the features you are generating.
    \item The \textbf{Relevant Dependency Structure} shows you exactly how the conditioning features relate to the features you need to generate.
    \item Use the \textbf{Example Data} to understand the format, range, and statistical properties of the real data.
    \item The generated table should contain columns for exactly the features listed in $\texttt{\{features\_to\_generate\}}$.
    \item Ensure the generated values are plausible and respect the learned dependencies.
\end{itemize}

\label{prompt:data_gen}
\end{tcolorbox}

\begin{tcolorbox}[
    float*=t, width=\textwidth, floatplacement=htbp,
    colback=purple!10!white,
    colframe=purple!75!black,
    title={Prompt for Independent Feature Synthesis ($\pi_\mathtt{iso}$)},
    fonttitle=\bfseries\color{white},
    arc=2mm,
    boxrule=1pt
]
\small

You are a helpful AI assistant that completes tabular data records by generating values for remaining features.

\vspace{\medskipamount}

\textbf{Task Description:}

The core, structurally-dependent features of a dataset have already been generated. 
Your task is to generate plausible values for the remaining isolated features, ensuring they are statistically consistent with the core features.

\vspace{\medskipamount}

\textbf{Conditioning Data (Generated Graph-Based Features):}
\begin{verbatim}
{graph_based_values}
\end{verbatim}
\vspace{\medskipamount}

\textbf{Features to Generate (Isolated Features):}
\begin{verbatim}
{isolated_features_to_generate}
\end{verbatim}
\vspace{\medskipamount}

\textbf{Example Data:}
\begin{verbatim}
{related_few_shot_data}
\end{verbatim}
\vspace{\medskipamount}

\textbf{Your Task is to Generate the Independent Values:}
\begin{itemize}[leftmargin=*, itemsep=2pt]
    \item Your primary task is to generate realistic data for the features listed in $\texttt{\{isolated\_features\_to\_generate}\}$.
    \item This generation must be \textbf{conditioned} on the complete set of core features provided in \textbf{Conditioning Data}.
    \item The features you are generating do not have direct parent-child dependencies in the learned graph, but their values should still be plausible and consistent in the context of the entire data record.
    \item Use the \textbf{Example Data} to understand the typical format and statistical properties of these isolated features.
    \item The generated table should contain columns for exactly the features listed in $\texttt{\{isolated\_features\_to\_generate}\}$.
\end{itemize}
\label{prompt:data_gen_iso}
\end{tcolorbox}

\clearpage

\end{document}